\documentclass{article}

\usepackage[final]{corl_2026}         %

\usepackage{amsmath, amsfonts, amssymb, amsthm, mathtools}
\DeclareMathAlphabet{\mathbbold}{U}{bbold}{m}{n}
\usepackage{bm}
\usepackage{graphicx}
\usepackage{booktabs}
\usepackage{multirow}
\usepackage{wrapfig}
\usepackage{algorithm}
\usepackage{algorithmic}
\usepackage{xcolor}                     %
\usepackage{xspace}
\usepackage{caption}                    %
\usepackage{cleveref}                   %
\usepackage{enumitem}                   %
\usepackage{amsmath}
\usepackage{titletoc}
\usepackage[svgnames]{xcolor} %
\usepackage[most]{tcolorbox}
\usepackage{subcaption}
\usepackage{bbm}

\makeatletter
\newcommand{\longdash}[1][2em]{%
  \makebox[#1]{$\m@th\smash-\mkern-7mu\cleaders\hbox{$\mkern-2mu\smash-\mkern-2mu$}\hfill\mkern-7mu\smash-$}}
\makeatother
\newcommand{\omitskip}{\kern-\arraycolsep}

\def\Acronym{\texttt{Q-OIL}}
\def\FullMethod{\textcolor{teal}{\textbf{Q}}-\textcolor{teal}{\textbf{O}}ptimistic
\textcolor{teal}{\textbf{I}}ntervention \textcolor{teal}{\textbf{L}}earning}

\usepackage{relsize}

\newcommand{\titlett}[1]{{\ttfamily\bfseries\larger[1] #1}}

\title{\titlett{Q-OIL}: Optimistic Exploration in Learning from Human Interventions}

\author{
\begin{tabular}{c}
Pranav Teegavarapu$^{1,*}$ \qquad
Sriyash Poddar$^{1,}$\thanks{Corresponding authors: \texttt{\{pranavnt,sriyash\}@cs.washington.edu}} \qquad
Rico Qi$^{1}$ \\[6pt]
Cleah Winston$^{1}$ \qquad
Krisha Khandelwal$^{1}$ \qquad
Aykut Onyol$^{2}$ \\[6pt]
Natasha Jaques$^{1,\ddagger}$ \qquad
Abhishek Gupta$^{1,\ddagger}$ \\[10pt]
{\normalfont\small
$^{1}$University of Washington
\qquad
$^{2}$Toyota Research Institute
} \\[5pt]
{\normalfont\footnotesize
$^{*}$Equal contribution
\qquad
$^{\ddagger}$Equal advising
} \\[3pt]
{\normalfont\footnotesize
Project website: \url{https://weirdlabuw.github.io/qoil}
}
\end{tabular}
}

\begin{document}
\maketitle
\vspace{-0.5cm}
\begin{abstract}
As robots are deployed in the real world, they will inevitably make mistakes. Human interventions provide valuable feedback for improving their behavior, enabling them to adapt online to new scenarios. However, studying human-in-the-loop robot learning is difficult: real-world experiments are slow and expensive, so most algorithm development relies on simulated human models. We show that common human modeling assumptions do not match how real people intervene in practice. Through analyzing intervention data from human participants, we find that humans tend to intervene when agent progress stalls, and that their corrective actions are often suboptimal. Motivated by these findings, we introduce a progress-based model of human intervention that achieves higher rank correlation with real-user algorithm rankings, enabling realistic algorithm benchmarking and development in simulation. Building on these insights, we propose \Acronym{}, a method that adds optimistic bonuses to intervention transitions in reinforcement learning and softly guides exploration toward human corrections. We find that decoupled critics are critical for stability because they prevent the bonuses from backing up to pre-intervention states.  As a result, \Acronym{} learns more efficiently than competitive baselines, achieving 10--50\% higher success in simulation and on real robots, and improving pretrained VLAs on two bimanual manipulation tasks. As we ground both evaluation and algorithms in realistic human behavior, this work provides a practical path toward scalable robot learning from interventions.
\end{abstract}

\keywords{Human-in-the-loop, Robot Learning, Reinforcement Learning}

\section{Introduction}
\label{sec:intro}

Robot policies trained offline on large-scale datasets are brittle~\citep{zhou2025libero}, often struggling to generalize beyond their training distributions~\cite{hu2024data} and failing across diverse testing conditions. For robust deployment, robots must adapt online from feedback~\citep{ouyang2022training, ross2011reduction, yin2025sgft, pi_star_06_2025}. As robots fail around human observers, we can rely on human feedback through corrections~\citep{michael2019hg, jiang2024transic, Liang2024LearningTL}, where humans intervene when the robot gets stuck or makes mistakes. Such signals arise organically in the real world; for example, users may physically steer a manipulator toward the goal~\citep{bajcsy2018learning,Mandlekar2020HumanintheLoopIL, pi_star_06_2025} or take over in self-driving systems~\cite{spencer2022expert}. Interventions can be viewed as more than resets or recovery actions~\cite{spencer2022expert}. Corrective actions convey implicit preference signals, i.e., information about failures in the robot's behavior and potential ways to improve~\cite{liu2022goal, tamer09knoxx, xie2022paint, luo2024rlifinteractiveimitationlearning, Korkmaz2025MILEMI, Lindner2022InteractivelyLP, spencer2021feedbackimitationlearningregimes, ainsworth2019mostatesmoproblems}.

Despite recent progress in robot learning from interventions, algorithm development remains difficult because real-world interaction is slow and expensive. Simulation offers a scalable alternative~\cite{jain2025polarisscalablerealtosimevaluations}. For intervention learning, we also require synthetic human models to determine when and how to intervene~\cite{luo2024rlifinteractiveimitationlearning}. These enable repeatable evaluation and systematically guide algorithmic components (e.g., how to incorporate corrections, the exploration-safety trade-off, robustness to noise) before deployment. However, common synthetic human models and learning algorithms are insufficiently grounded in real-world behavior. They make assumptions that we show are \emph{not reflective of real human intervention data}. This creates a gap between methods trained on synthetic human models and their performance when exposed to real users. We need a principled understanding of \emph{when} and \emph{how} humans intervene, both to guide our algorithmic design choices and to evaluate these methods. We first analyze real human interventions in robot manipulation. Our analysis suggests both how to better model human interventions and how to learn from them.%

\begin{figure*}
    \centering
    \includegraphics[width=\linewidth]{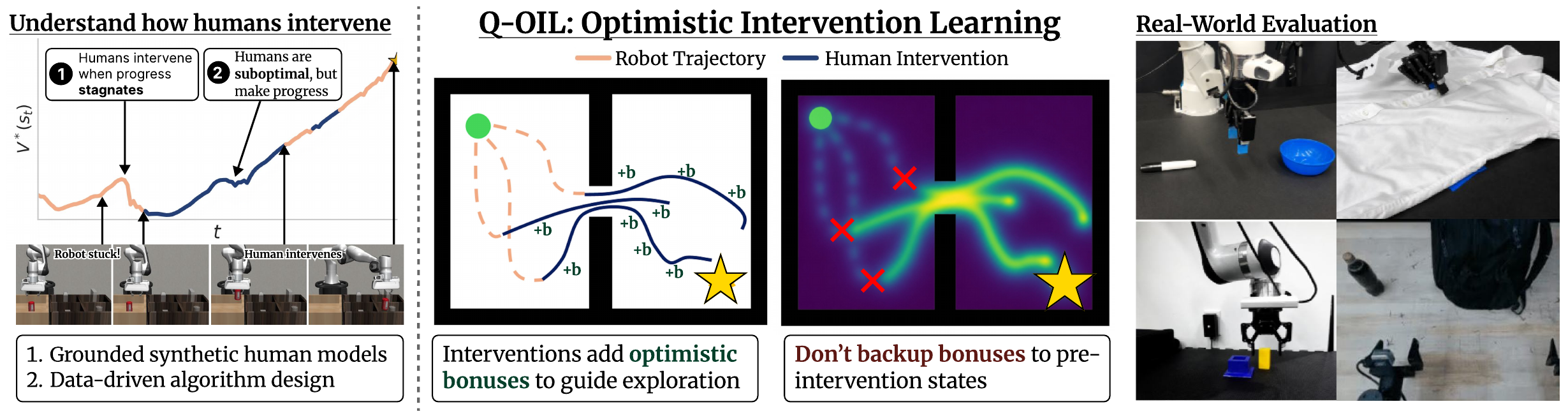}
    \caption{\footnotesize Overview of \Acronym{}. Real-user data shows that humans intervene when robot progress stalls and provide suboptimal corrections. \Acronym{} treats interventions as optimistic exploration bonuses while preventing bonuses from propagating to non-intervention states for faster real-world learning.}
    \label{fig:teaser}
    \vspace{-0.8cm}
\end{figure*}

First, we find that rather than reacting to instantaneous suboptimality, humans intervene when the robot's progress stagnates over a horizon. This pattern aligns with cognitive models of caregiving: humans act on the agent’s ability over time rather than on isolated states~\citep{shachnai2025bayesians}. Given this finding, we propose a progress-based synthetic intervention model that accounts for value improvement over a horizon, better matches observed humans, and yields algorithm rankings more consistent with real human evaluation (Fig.~\ref{fig:correlation}). Second, this analysis shows that real corrections are suboptimal and can diverge from the learner policy during training, which suggests that algorithms for learning from interventions must be robust to noisy, divergent interventions, i.e., use feedback to adapt robot behavior without compromising asymptotic performance. This motivates our key algorithmic insight: use interventions as optimistic guidance for exploration, rather than optimal supervision. This speeds up training without compromising asymptotic performance.  

Building on these insights, we introduce \textbf{\Acronym{}} (\FullMethod{}), a simple online RL framework that learns from real human corrections and efficiently handles diverse users and suboptimal interventions.
We start by providing an additional positive reward, i.e., an optimism bonus, to intervention transitions.
Intuitively, the optimism bonus uses interventions to guide exploration toward potentially successful behaviors.
 The influence of optimism naturally decays as interventions become less frequent, shifting learning back to the original task reward. To make this stable, we introduce a critical decoupling of optimism from TD learning~\citep{Sutton1988TD}: the optimistic critic drives exploration and policy updates, while a task-only critic provides TD targets, preventing catastrophic bonus propagation to non-intervention states. Importantly, \Acronym{} is agnostic to the underlying off-policy RL algorithm, allowing it to be incorporated into different policy learning frameworks. We demonstrate this across simulated and real-world robot manipulation tasks with human participants, and further scale \Acronym{} to improve pretrained vision-language-action (VLA) policies on challenging manipulation tasks.

In summary, we make the following contributions:
\begin{enumerate}[topsep=0pt, itemsep=-0.75ex, partopsep=1ex, parsep=1ex, leftmargin=0.5cm]
    \item We analyze real data to ground synthetic human models for realistic simulation evaluation and informed algorithm design. Our findings show that interventions occur when robot progress stagnates, motivating a progress-based model that promotes improved experiment design.
    \item Based on human analysis, we propose \Acronym{}, a simple algorithm that treats interventions as optimistic guidance for exploration rather than optimal supervision targets.
    \item Across diverse real and simulated robot manipulation tasks with human participants, \Acronym{} learns faster, reduces human effort, and remains robust to suboptimal and diverse interventions. We further scale it to improve pretrained VLAs on two bimanual manipulation tasks.
\end{enumerate}

\section{\vspace{-0.2cm}Related Work}\label{sec:related_work}

\textbf{Learning from Human Demonstrations.} Behavior cloning from offline demonstrations~\citep{Argall-2009-17073, osa18imitation, memmel2025strap} is a standard approach for training robot policies, but suffers from compounding errors under distribution shift at deployment~\citep{ross2011reduction, spencer2021feedbackimitationlearningregimes}. Reinforcement learning can learn robust behaviors from task rewards~\citep{haarnoja2018soft, schulman17ppo}, but is often sample-inefficient due to undirected exploration. Recent works~\citep{Rajeswaran-RSS-18, nair2021awac, kostrikov2022offline, yin2025sgft, ball2023efficient} warm-start RL with offline datasets, and particularly, \citet{Lu2022ImitationIN} combine behavior cloning with off-policy actor-critic updates. However, collecting task demonstrations is costly. In contrast, we leverage easy-to-provide online interventions~\citep{michael2019hg} to bias policy behavior during training. 

\textbf{Interactive Learning.} Interactive learning methods~\citep{ross2011reduction, michael2019hg, jiang2024transic, luo2024hilserl} address the drawbacks of offline approaches by collecting online feedback, reducing compounding error from quadratic to linear regret~\citep{ross2011reduction}. A natural class of methods allows humans to override policy behavior with corrections~\citep{michael2019hg, Mandlekar2020HumanintheLoopIL, xie2022paint, luo2024hilserl}, which prior work uses for behavior-cloning~\citep{michael2019hg, Mandlekar2020HumanintheLoopIL}, as rewards~\citep{bajcsy2018learning}, or for reinforcement learning~\citep{luo2024hilserl}. Notably, \citet{luo2024hilserl} incorporate interventions into the replay buffer, making minimal assumptions about human intervention behavior, and achieve strong real-world results. Beyond corrective actions, interventions also provide implicit feedback: RLIF~\citep{luo2024rlifinteractiveimitationlearning} uses them as implicit negative feedback while \citet{spencer2022expert} uses interventions to constrain the learner's value function.
Our work investigates how corrections can guide exploration. Inspired by intrinsic-reward-driven exploration~\citep{brafman2002rmax, pathakICMl17curiosity}, \Acronym{} adds optimistic bonuses to critic targets for intervention transitions. Prior exploration methods use self-supervised prediction errors~\citep{pathakICMl17curiosity} or explicit human feedback~\citep{torne2023breadcrumbs}, while we use naturally occurring corrective interventions to guide exploration.

\textbf{Modeling Human Behavior.} Learning from interventions requires accurate models of when and how humans intervene. Because real-world training is costly~\citep{yin2025sgft}, prior work relies on approximate human models: TAMER~\citep{tamer09knoxx} learns a scalar reward predictor, while others use Boltzmann-rational models~\citep{bajcsy2018learning, biyik2022learning}, probit triggers~\citep{Korkmaz2025MILEMI}, or threshold-based rules tied to instantaneous suboptimality~\citep{luo2024rlifinteractiveimitationlearning}. A common assumption is that human corrections are Markovian, i.e., merely reactions to instantaneous suboptimality, which our analysis contradicts. Consistent with cognitive theories of utility trade-offs~\citep{kahneman1979prospect} and computational models of caregiving~\citep{shachnai2025bayesians}, our proposed model accounts for agent progress over a short horizon when triggering interventions, grounded in real human data.

\section{Problem Statement}\label{sec:prob}

We build on the framework of interactive imitation learning, focusing on learning from \emph{human interventions}~\citep{michael2019hg}. Consider an interactive control setting over an MDP $\mathcal{M}=(\mathcal{S},\mathcal{A},\mathcal{T}, \mathcal{R}, \gamma,\rho_0)$, where 
$\mathcal{S}$ is the state space, $\mathcal{A}$ is the action space, and $\mathcal{T},\mathcal{R},\rho_0, \gamma$ are the transition kernel, (sparse) reward function, initial state distribution, and discount, respectively.
The objective is to learn a policy $\pi_\theta(a|s)$ that maximizes expected discounted return, i.e., $\pi_\theta=\arg\max_\theta \ \mathbb{E}_{\pi_\theta}\big[\sum_{t\ge 0}\gamma^t r(s_t)\big]$. 
During deployment, a human observes the agent and may intervene. We model the human as a decision function $\mathcal{H}=(g,\pi_h)$ with two components: a \emph{gating function} $g$ that decides \emph{when} to intervene, and a \emph{human policy} $\pi_h$ that decides \emph{how} to intervene. To allow temporal context, both may depend on a recent history $\tau_{t-L:t}=(s_{t-L},a_{t-L},\dots,s_t)$, so, $g:\mathcal{S}^{L+1}\times\mathcal{A}^{L}\to[0,1]$ outputs the probability of intervening at time $t$, and $\pi_h:\mathcal{S}^{L+1}\times\mathcal{A}^{L}\to \mathcal{A}$ returns a distribution over actions. The resulting rollout policy is the mixture
$
\pi'(a\mid s_t,\tau_{t-L:t}) \;=\; g(\tau_{t-L:t})\,\pi_h(a\mid \tau_{t-L:t}) \;+\; \big(1-g(\tau_{t-L:t})\big)\,\pi_\theta(a\mid s_t)
$. Thus, modeling human behavior requires simulating $g$ and $\pi_h$. In the following sections, we (i) collect and analyze real human interventions to characterize when and how people intervene (Section~\ref{sec:rethinking}), (ii) propose a non-Markovian progress-based model for $g$ validated against this data (Section~\ref{sec:pam}), and (iii) leverage this analysis to develop a new algorithm addressing key failure modes of prior approaches in learning from interventions (Section~\ref{sec:rl_method}).

\vspace{-0.1cm}
\section{Rethinking Simulated Interventions}
\label{sec:rethinking}
Developing methods that learn from interventions requires extensive experiments, but real-user studies are slow to reproduce, making it impractical to explore the full design space in deployment. Thus, researchers rely on simulated human models for scalable training and evaluation to build better learning methods. However, if the human model is misaligned with reality, methods optimized for these models and algorithmic choices inferred from such experiments may not transfer to learning from real users. This motivates a central question: \emph{when and how do humans actually intervene?}

\textbf{Prior Approaches to Simulating Interventions.}\label{sec:prior-human-models} Prior works build simulated models of both intervention timing and corrective actions, with different assumptions. \citet{Korkmaz2025MILEMI} model intervention decisions using a probit choice model that accounts for a nominal human action and the human's model of the robot. \citet{luo2024rlifinteractiveimitationlearning} uses a Markovian model of intervention decisions based on instantaneous action suboptimality relative to a reference policy. Beyond deciding when to intervene, simulated interventions require a corrective policy $\pi_h$, which may be instantiated by a reference policy $\pi^*$ or a weaker policy to model imperfect corrections. Below, we analyze human interventions and find that humans intervene when the robot's perceived progress is slow and that human corrections are suboptimal. These findings contradict prior assumptions, and thus, conclusions based on such models can diverge from experiments with real humans.

\textbf{Collecting Real Human Intervention Data.}\label{sec:intv_data}
We ran a pilot study with 8 participants supervising a simulated robot across 3 manipulation tasks, yielding 24 runs and $\sim200$k transitions. Participants provided interventions during training (Section~\ref{sec:experiments}), following \citet{luo2024hilserl}. We recorded when takeovers occurred and how users corrected the robot across learning stages. The study received IRB approval (Appendix~\ref{app:human_exps_analysis}). For analysis, we trained task-specific reference policies and value functions, $\pi^*$, $Q^*(s,a)$, and $V^*(s)$, using RLPD~\citep{ball2023efficient} with sparse rewards, allowing us to quantify interventions in terms of optimality and convergence. We include participant scores in Appendix~\ref{app:human_exps}.

\subsection{Analyzing Real Human Intervention Data}\label{sec:human_analysis} 
\begin{tcolorbox}[
    colback=blue!10,
    colframe=blue!50,
    arc=4pt,
    boxrule=0pt,
    left=6pt,right=6pt,top=6pt,bottom=6pt
]
Finding 1: Human intervention gating is non-Markovian.
\end{tcolorbox}
\vspace{-0.1cm}

\begin{wrapfigure}{r}{0.5\linewidth}
    \vspace{-0.5cm}
    \centering
    \includegraphics[width=\linewidth]{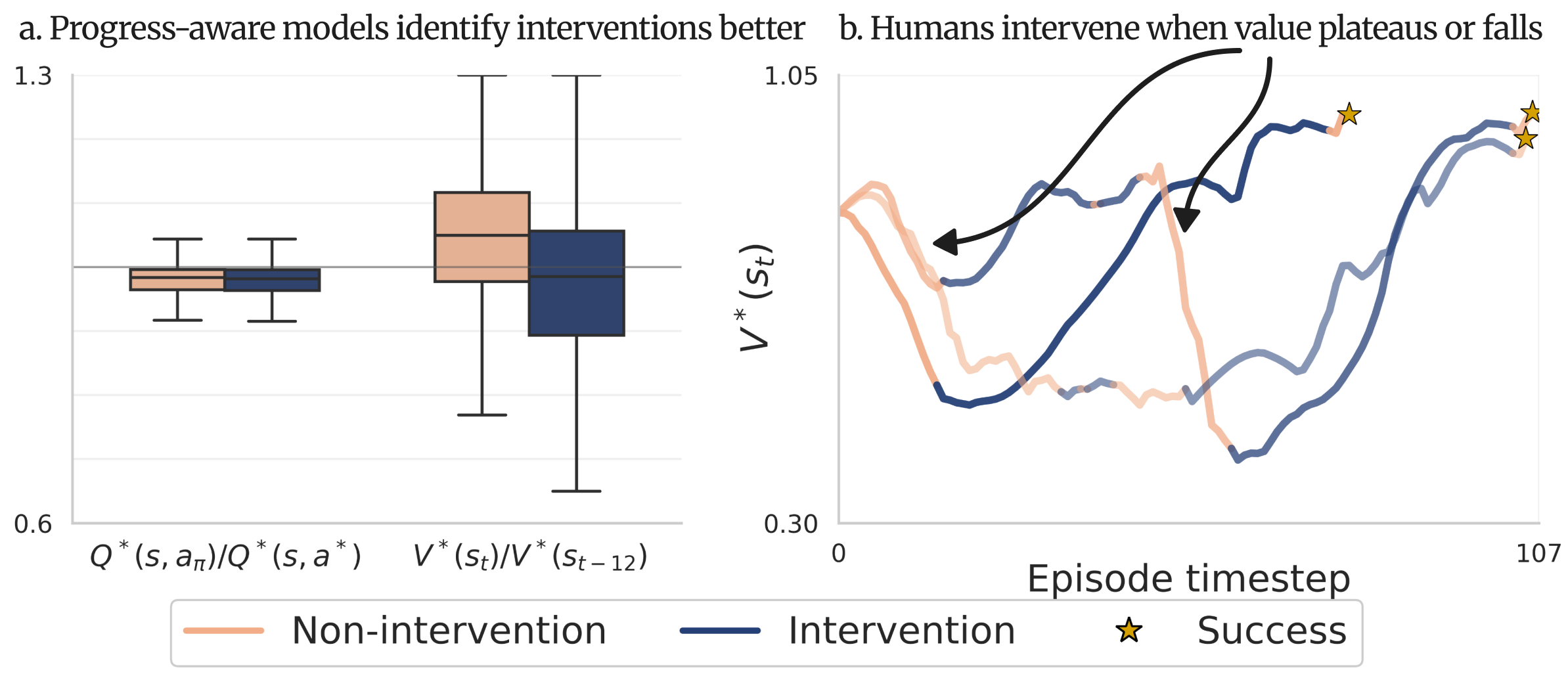}
    \caption{\footnotesize (a) Progress separates interventions better than instantaneous values. (b) Qualitatively, humans intervene when value plateaus or declines.
    \label{fig:action_subopt}}
    \vspace{-0.25cm}
\end{wrapfigure}

Typically, the gating function is modeled as \emph{Markovian} and reactive to the suboptimality of the action, implying that intervention timing depends only on the current state and action. Figure~\ref{fig:action_subopt}a contradicts this: $Q^*(s, a_\pi)/Q^*(s, a^*)$ is nearly identical for intervention vs.\ non-intervention states, indicating similar instantaneous values. But, when we measure $k$ step progress as $V^*(s_t)-V^*(s_{t-k})$, the three separate value trajectories (Fig.~\ref{fig:action_subopt}b) show that
takeovers follow plateaus or drops in $V^*(s_t)$, suggesting stagnation rather than single-step failure. This matches evidence that corrections can lag agent behavior~\citep{tamer09knoxx}. Thus, Markovian models insufficiently represent humans, leading to experimentally invalid conclusions.

\begin{tcolorbox}[
    colback=blue!10,
    colframe=blue!50,
    arc=4pt,
    boxrule=0pt,
    left=6pt,right=6pt,top=6pt,bottom=6pt
]

Finding 2: Human interventions are suboptimal demonstrations of the target behavior.
\end{tcolorbox}
\vspace{-0.2cm}

\begin{wrapfigure}[7]{r}{0.25\linewidth}
    \vspace{-0.6cm}
    \centering
    \includegraphics[width=\linewidth]{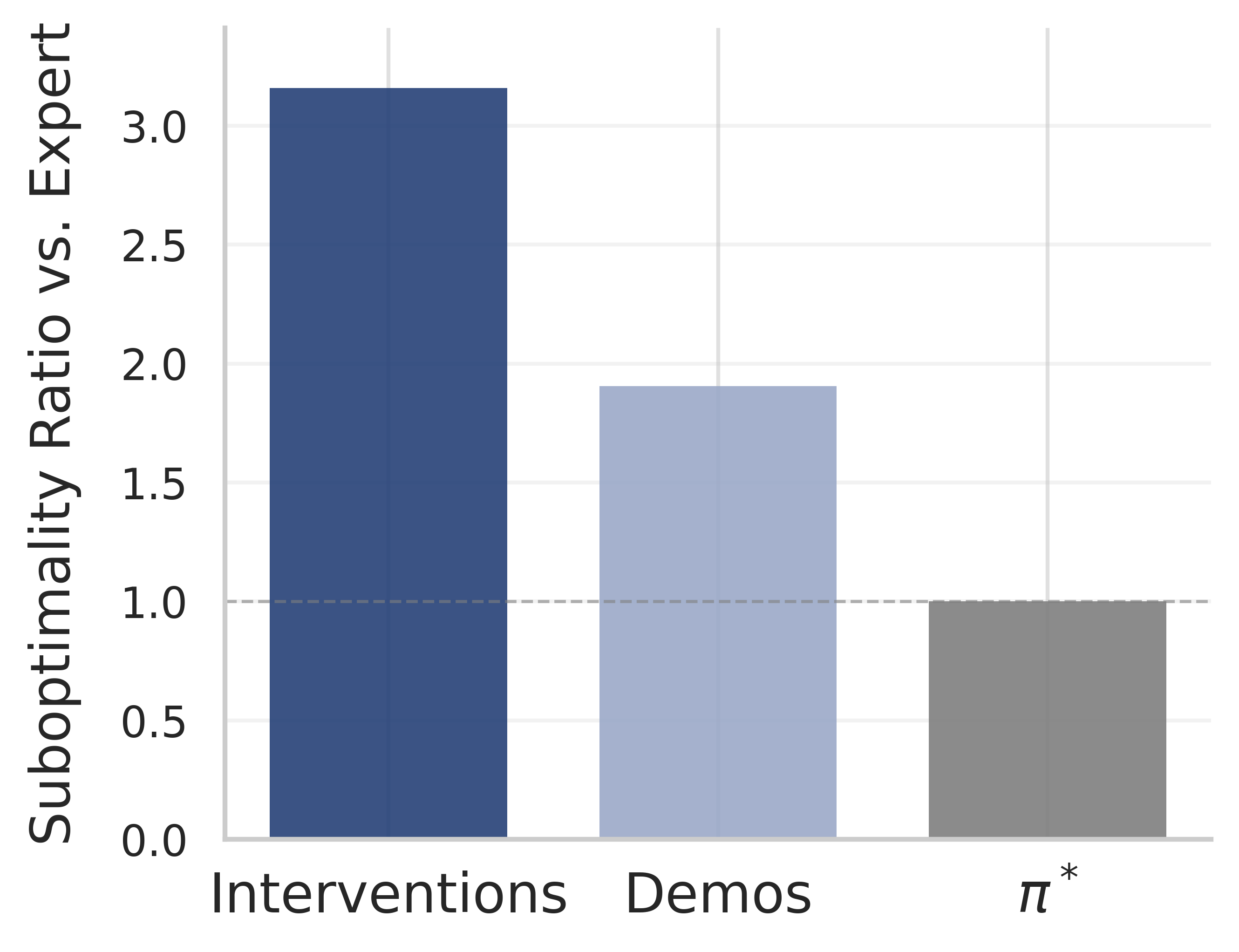}
    \caption{\footnotesize Interventions are suboptimal to demos and $\pi^*$.\label{fig:subopt}}
\end{wrapfigure}
Several methods~\cite{michael2019hg, Mandlekar2020HumanintheLoopIL} treat human corrections as demonstrations, assuming $\pi_h$ is near-optimal. To quantify takeover quality, we compare the value of the intervention action to that of the expert action at the same state, and compute a suboptimality score defined in Appendix~\ref{app:progress_metric}. Under this metric, interventions are more suboptimal than $\pi^*$ and human demonstrations. Figure~\ref{fig:action_subopt} shows that interventions do not consistently increase $V^*(s_t)$. These results motivate methods that are robust to noisy corrections.

\par
\WFclear

\begin{tcolorbox}[
    colback=blue!10,
    colframe=blue!50,
    arc=4pt,
    boxrule=0pt,
    left=6pt,right=6pt,top=6pt,bottom=6pt
]
Finding 3: Human corrections diverge from the learned policy over training.
\end{tcolorbox}
\begin{wrapfigure}{l}{0.16\linewidth}
    \centering
    \includegraphics[width=\linewidth]{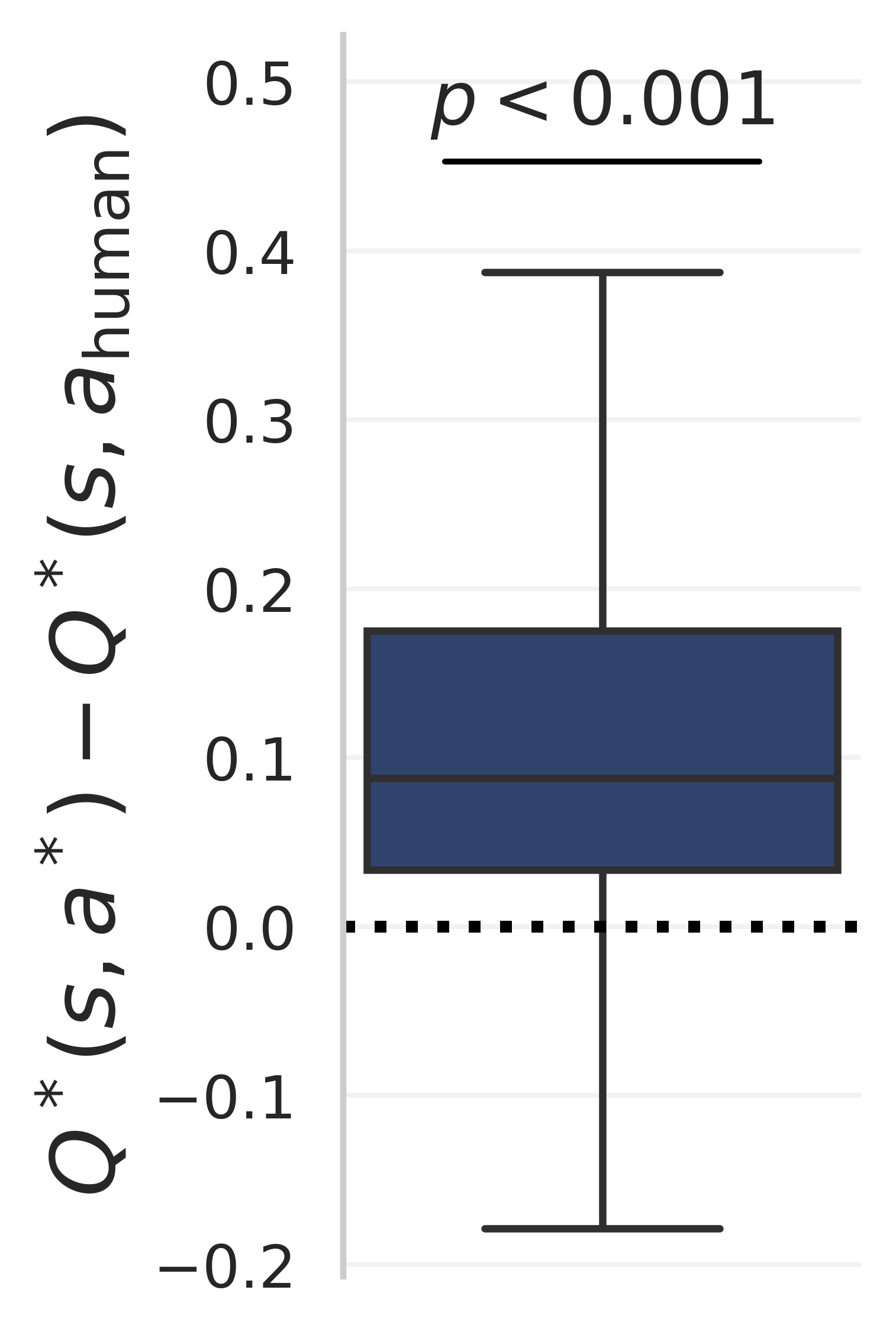}
\end{wrapfigure}
Across the intervention dataset ($\sim40$k intervention transitions), we find a systematic mismatch between human corrections and the learner's converged behavior. Using $Q^*$, we compare values between the
converged policy ($a^\text{final}$) and human actions and observe  $Q^*(s,a^\text{final}) \geq Q^*(s, a_\text{human})$, implying that at the same states, the final policy prefers different actions with higher values. As both have high success ($\sim$70-80\%), the gap is better explained as divergence in solution modes, which is a result of sparse rewards admitting multiple solutions (e.g., left- vs.\ right-side grasps). Human supervisors do not observe the full policy, so may provide corrections with an alternative strategy. Therefore, treating corrections as strict supervision can induce undesirable multi-modality, so instead we motivate using interventions as soft guidance for exploration to speed up training.
\par

\subsection{A Non-Markovian Gating Model for Human Interventions}
\label{sec:pam}
Motivated by these findings, we define $g$ as a progress-aware model (PAM) of when humans intervene over recent history $\tau_{t-k:t}$:

\vspace{-0.6cm}
\begin{equation}
\begin{aligned}
g(\tau_{t-k:t})
&= \Pr(\nu_t=1 \mid \tau_{t-k:t})
=
\begin{cases}
\alpha, & \text{if } V^*(s_t) - V^*(s_{t-k}) < \delta,\\
\beta,  & \text{otherwise}.
\end{cases}
\end{aligned}
\end{equation}
\vspace{-0.6cm}

using $V^*(s_t)-V^*(s_{t-k})$ as $k$-step progress, and where $k$ is the horizon, $\delta$ the progress threshold, and $\alpha,\beta$ capture intervention stochasticity. Thus, interventions are likely when recent value progress stagnates (including failures where $V^*(s_t)-V^*(s_{t-k})<0$). We tune $k,\delta$ by grid search (Appendix~\ref{app:intervention_models}) on real human data. To model how humans intervene, i.e., $\pi_h$, we use an intermediate RL checkpoint ($\sim$50--70\% success), reflecting that corrections are imperfect.

\begin{wrapfigure}{l}{0.3\linewidth}
  \vspace{-0.6cm}
  \centering
  \includegraphics[width=0.9\linewidth]{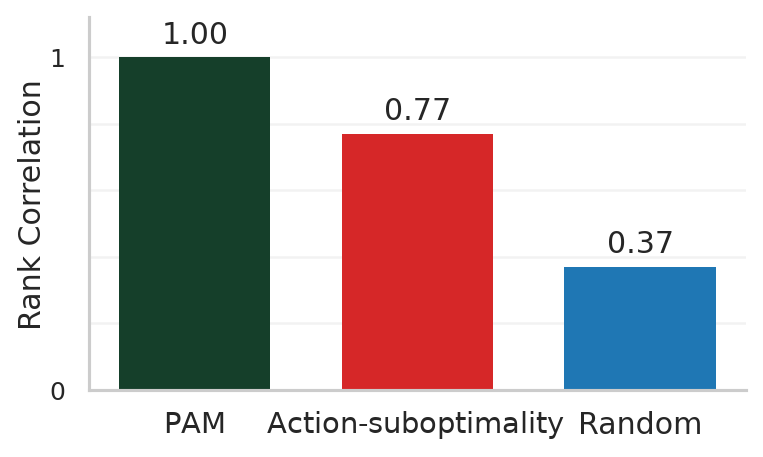}
  \caption{\footnotesize $\rho$ between approximate models and real human algorithm rankings.}
    \label{fig:correlation}
  \vspace{-0.5cm}
\end{wrapfigure}

\textbf{Validation Against Real-Human Rankings.}
A useful intervention model should predict which algorithms succeed with real users. We validate $g$ by held-out intervention prediction and ranking agreement with real-human evaluations. Across all runs, we rank six learning algorithms (Section~\ref{sec:experiments}) by final success under three models: our PAM model, action suboptimality~\citep{luo2024rlifinteractiveimitationlearning}, and a random model calibrated to the true intervention rate, and compare these rankings to human evaluations using Spearman correlation.

\textbf{Our progress-based gating approach is a stronger approximate human model for evaluating robot intervention learning.} Figure~\ref{fig:correlation} shows stronger agreement with real-human rankings ($\rho=1.0$) than alternative gating models, providing a more reliable proxy for benchmarking algorithms with real users.
We provide further discussion in Appendix~\ref{app:intervention_models}.

\section{How Should We Use Human Interventions for Robot Policy Learning?}
\label{sec:rl_method}

As outlined in Section~\ref{sec:human_analysis}, prior methods for intervention learning rely on assumptions unsupported by real data. Given this, we ask: \emph{how should we use imperfect human interventions for robots?}
\vspace{-0.2cm}
\begin{tcolorbox}[
    colback=blue!10,
    colframe=blue!50,
    arc=4pt,
    boxrule=0pt,
    left=6pt,right=6pt,top=6pt,bottom=6pt
]
Human interventions provide optimistic signals to guide exploration during RL.
\end{tcolorbox}
\begin{wrapfigure}{r}{0.33\textwidth}
    \centering
    \includegraphics[width=\linewidth]{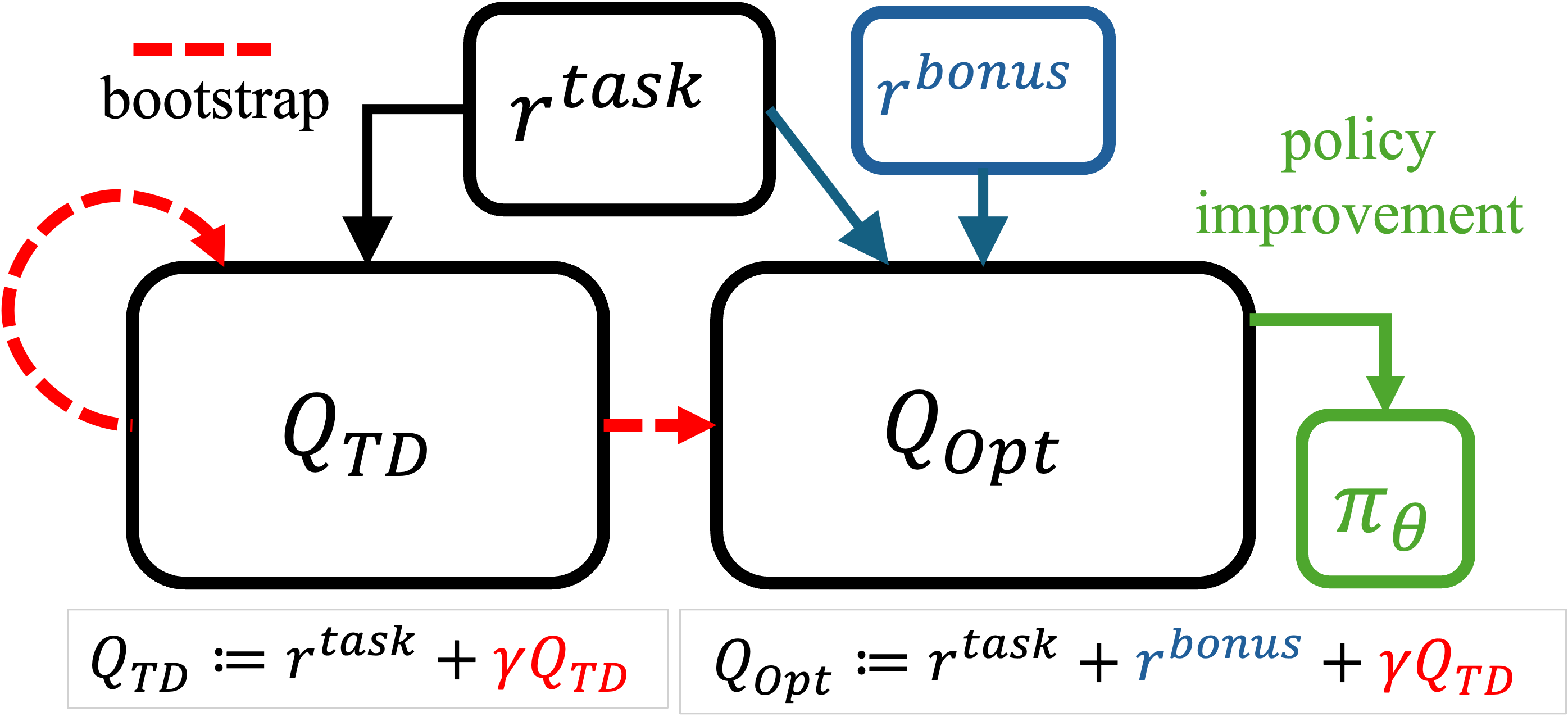}
    \caption{\footnotesize Overview of \Acronym{}\label{fig:method}}
    \vspace{-0.5cm}
\end{wrapfigure}
To this end, grounded in our findings about human behavior, we present an approach based on off-policy model-free RL and optimistic exploration~\citep{brafman2002rmax}: \Acronym{}, which adds a positive reward bonus at intervention transitions, encouraging exploration informed by corrections without forcing
strong alignment to potentially noisy and divergent corrections (Findings 2 and 3). Standard off-policy RL learns a critic by using the given rewards and bootstrapping the value of future states. Training a single critic with the optimism bonus and bootstrapped updates would propagate this bonus backward, reinforcing the actions that triggered the intervention. We address this catastrophic over-optimism via decoupling the critic into an optimistic and task critic, as illustrated in Fig.~\ref{fig:method}. This is critical because an optimistic critic (trained on task reward + bonus) bootstraps from the task critic (trained only on task reward), thereby confining the bonus to intervention transitions. The optimistic critic improves the policy, while the task critic is an unbiased estimate of the true value function. We discuss this further in Appendix~\ref{app:toy_decoupled_q}. Finally, we show that this human-informed method improves performance with real users and the better-aligned synthetic human model.

\subsection{\FullMethod{}}
\begin{wrapfigure}{r}{0.47\textwidth}
    \vspace{-1.6cm}
  \begin{minipage}{\linewidth}
    \begin{algorithm}[H]
    \caption{\footnotesize}
    \label{alg:method}
    \begin{algorithmic}[1]
    \footnotesize
    \setlength{\itemsep}{0pt}
    \setlength{\parskip}{0pt}
    \REQUIRE $\pi_\theta$, $\pi^{\text{int}}$, $D_\pi$, $D_\text{int}$, bonus $b > 0$
    \FOR{trial $i = 1$ to $N$}
        \STATE \texttt{\textcolor{NavyBlue}{// Training decoupled critics}}
        \STATE Train $Q_{\text{TD}}$ on $D_\pi$ via $\mathcal{L}_{\text{TD}}$
        \STATE Train $Q_{\text{Opt}}$ on $D_\pi$ via $\mathcal{L}_{\text{Opt}}$
        \STATE \texttt{\textcolor{NavyBlue}{// Policy improvement}}
        \STATE Update $\pi_\theta$ on $D_\pi$ and $D_\text{int}$ via $\mathcal{L}_{\theta}$
        \FOR{timestep $t = 1$ to $T$}
            \STATE // \texttt{\textcolor{NavyBlue}{rollout}} $\pi_\theta$
            \IF{$\pi^{\text{int}}$ intervened at $t-1$}
                \STATE $r^\text{bonus}_{t-1} = b$
                \STATE append $(s_{t-1}, a^{\text{int}}_{t-1})$ to $D_\text{int}$
            \ELSE
                \STATE $r^\text{bonus}_{t-1} = 0$
            \ENDIF
            \STATE add $(s_{t-1}, a_{t-1}, r^{\text{task}}_{t-1}, r^\text{bonus}_{t-1}, s_t)$ to $D_\pi$
        \ENDFOR
    \ENDFOR
    \end{algorithmic}
    \end{algorithm}
  \end{minipage}
\end{wrapfigure}

Practically, we learn a policy $\pi_\theta$, task critic $Q_\text{TD}$, and optimistic critic $Q_\text{Opt}$ using an online off-policy actor–critic algorithm~\citep{haarnoja2018soft, ball2023efficient}, though in principle it may improve any RL method. In Algorithm~\ref{alg:method}, $\pi_\theta$ is rolled out to interact with the environment, while a human or an approximate model supervisor intervenes with corrective actions. $D_\pi$ stores transitions $(s_t, a_t, r^{\text{task}}_t, s_{t+1})$ and bonus reward $r^\text{bonus}_t=b*\mathbbold{1}(\text{intervene})$, while $D_{\text{int}}$ stores intervention state-action pairs $(s_t, a_t)$. Optionally, we can initialize $D_\pi$ and $D_\text{int}$ with a few offline demonstrations to warm-start training~\cite{ball2023efficient}. The backup critic $Q_\text{TD}$ is updated on $D_\pi$ by minimizing the standard Bellman regression loss,
$
\mathcal{L}_\text{TD} =
\mathop{\mathbb{E}}_{(s,a,r^{\text{task}}_t,s')\sim D_\pi}
\left[
Q_{\text{TD}}(s,a)
-
\left(
r^{\text{task}}_t+\gamma
\mathop{\mathbb{E}}_{a'\sim\pi_\theta(\cdot|s')}
Q_{\text{TD}}(s',a')
\right)
\right]^2
$. For the optimistic critic $Q_\text{Opt}$, we add the optimistic reward $r^\text{bonus}$ but use TD targets from $Q_\text{TD}$, while the actor $\pi_\theta$ objective combines (i) RL policy improvement using $Q_\text{Opt}$ and (ii) maximum likelihood on $D_{\text{int}}$:
\vspace{-0.2cm}
\[
\mathcal{L}_\text{Opt} =
\mathop{\mathbb{E}}_{(s,a,r^{\text{task}}_t,r^\text{bonus},s')\sim D_\pi}
\left[
Q_{\text{Opt}}(s,a)
-
\left(
\underbrace{r^{\text{task}}_t+\textcolor{red}{r^\text{bonus}}}_{\text{adding optimism}}
+
\gamma
\underbrace{
\mathop{\mathbb{E}}_{a'\sim\pi_\theta(\cdot|s')}
Q_{\textcolor{red}{\text{TD}}}(s',a')
}_{\text{using TD-backup}}
\right)
\right]^2
\]
\vspace{-0.1cm}
\vspace{-0.2cm}
\[
\mathcal{L}_\theta
=
-
\underbrace{
\mathop{\mathbb{E}}_{\substack{
s\sim D_\pi,
a \sim \pi_\theta(\cdot|s)
}}
\left[
Q_{\textcolor{red}{\text{Opt}}}(s,a)
\right]
}_{\text{optimistic policy improvement}}
-
\underbrace{
\mathop{\mathbb{E}}_{\substack{
(s,a)\sim D_{\text{int}}
}}
\left[
\lambda\log \pi_\theta(a|s)
\right]
}_{\text{BC-regularization}}
\]

In \Acronym{}, we apply the bonus to interventions, and during training we sample the replay uniformly. As a result, the probability of sampling bonus transitions is proportional to the cumulative intervention rate $\rho_t$. At convergence $\rho_t \to 0$, so $Q_{\text{Opt}} \to Q_{\text{TD}}$, i.e., the bonus anneals implicitly, leading to no asymptotic bias from the optimistic critics. Thus, \Acronym{} speeds up training while remaining robust to potential bias from human intervention. Additionally, \Acronym{} is a simple extension on top of any RL fine-tuning algorithm, and we extend this method to fine-tuning pretrained vision-language-action models (VLAs) in Algorithm~\ref{alg:vla_qoil}.

\textbf{BC regularization.} Interventions are suboptimal (Section~\ref{sec:human_analysis}) but provide a useful early directional signal compared to random exploration (see value improvement over corrections in Figure~\ref{fig:action_subopt}). We therefore regularize the actor with a BC term on intervention actions in $D_\text{int}$ with a fixed $\lambda \approx 0.1$.

\section{Experiments}
\label{sec:experiments}

Our experiments aim to answer whether \Acronym{} improves the ability to learn from interventions:
\begin{enumerate}[topsep=0pt, itemsep=-0.75ex, partopsep=1ex, parsep=1ex, leftmargin=0.5cm]
    \item in simulated robot experiments against the proposed model of human behavior;
    \item with real humans intervening on simulated robots;
    \item with real human interventions on a robot; and
    \item when fine-tuning pretrained VLAs on bimanual tasks.
\end{enumerate}
Finally, ablations investigate how components such as the decoupled critics and BC-regularization affect performance.

\subsection{Experimental Setup}

\begin{figure*}[t]
    \centering
    \includegraphics[width=\textwidth]{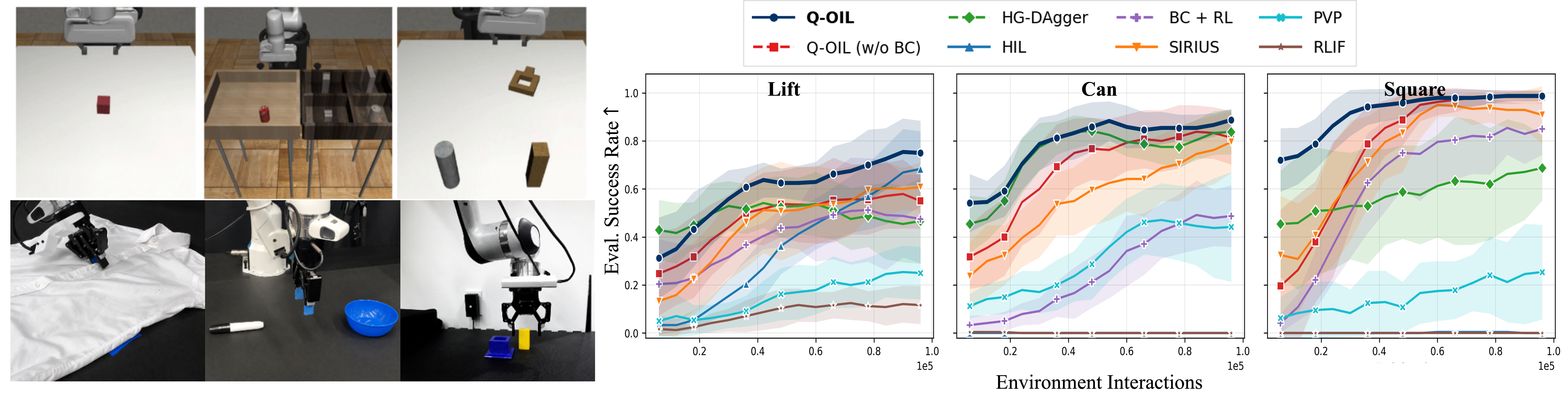}
    \caption{\footnotesize 
    \textbf{Left:} Evaluation tasks in simulation and on the Franka robot. 
    \textbf{Right:} On Robomimic with the progress-aware human model (PAM), \Acronym{} improves success rate and learning efficiency over baselines.
    }
    \label{fig:tasks_and_sim_results}
\end{figure*}

\textbf{Tasks.} We evaluate our method on real and simulated manipulation tasks using both simulated human models and real participants: (i) \textbf{Pen-in-Bowl:} grasp a pen and place it in a bowl; (ii) \textbf{Unbutton-Shirt:} open a shirt button, requiring fine-grained deformable-object manipulation; (iii) \textbf{Peg-Insertion:} pick and precisely insert a randomly placed peg; (iv) \textbf{Bag Assembly:} use two arms to pick up a bottle and insert it into a backpack's side pocket; (v) \textbf{Coffee Pouring:} mimicking a wet-lab task, pick a test tube filled with coffee beans from a rack and pour the beans into a beaker, requiring precise grasping and pouring; and (vi) \textbf{Simulated Robomimic:} three simulated pick-and-place/assembly tasks \citep{mandlekar2021matters}.

\textbf{Baselines.} We compare against methods spanning the three categories in Section~\ref{sec:related_work}: (i) \textbf{HIL}~\cite{luo2024hilserl}, which accelerates off-policy RL with online interventions in the replay buffer; (ii) \textbf{HG-DAgger}~\cite{michael2019hg}, a DAgger~\cite{ross2011reduction} variant that treats interventions as BC targets; (iii) \textbf{SIRIUS}~\cite{liu2022robot}, which applies weighted BC to interventions and on-policy data; (iv) \textbf{PVP}~\cite{Peng2023PVP}, which modifies the critic loss to prefer human actions; (v) \textbf{RLIF}~\cite{luo2024rlifinteractiveimitationlearning}, which assigns negative reward to pre-intervention states during off-policy RL; and (vi) \textbf{BC+RL}~\citep{Fujimoto2021TD3BC}, which combines BC on intervention actions with off-policy RL.

\textbf{Training Details.} We use a JAX-based implementation built on RLPD~\cite{ball2023efficient}. In simulation, we use our synthetic model PAM for reliable benchmarking and report success rates over 100k environment interactions. We then test algorithms with four real participants on two state-based simulated tasks. We narrow the reset distribution when real humans intervene in simulation to reduce training time. For real-robot tasks, we train RGB policies with wrist-mounted and third-person cameras on the Franka Panda robot and run training with a single human across three tasks. We warm-start experiments with 10-20 demonstrations, and evaluate each checkpoint for 10--20 episodes (100 for simulated tasks). For all our Franka experiments, we use RLPD~\cite{ball2023efficient} as our base algorithm. For our bimanual experiments, we fine-tune a pretrained VLA with EXPO-FT~\cite{dong2026expo}, which effectively learns a residual policy via interventions (See Appendix~\ref{app:vla_setup}). We include further details about the robot setup, tasks, and hyperparameters in Appendices~\ref{app:robot_setup} and \ref{app:implementation}.

\vspace{-0.2cm}
\subsection{Results}

\textbf{Does \Acronym{} improve success rate and learning efficiency in simulated tasks?}

In Figure~\ref{fig:tasks_and_sim_results} (right), we evaluate \Acronym{} on Robomimic using our progress-aware human model (PAM), which we show (Section~\ref{sec:pam}) better aligns with real human interventions and evaluations. Across tasks, \Acronym{} consistently achieves higher success rates with fewer interactions than baselines. This highlights that optimistic exploration with noisy corrections is more effective than directly imitating, penalizing, or naively adding interventions to the replay buffer.
We also compare all methods under existing human models and ablate expert quality in Appendices~\ref{app:intervention_models} and~\ref{app:perfect_experts}.

\textbf{Does \Acronym{} improve task performance and reduce \underline{real human} effort when participants interact with simulated robots?}

To evaluate performance under real human interventions, we recruited four participants to intervene on two Robomimic tasks across six methods, yielding 48 runs; see Appendix~\ref{app:human_exps_method} for details. 
In 
\begin{wrapfigure}{r}{0.5\textwidth}
    \vspace{-0.5cm}
    \centering
    \includegraphics[width=\linewidth]{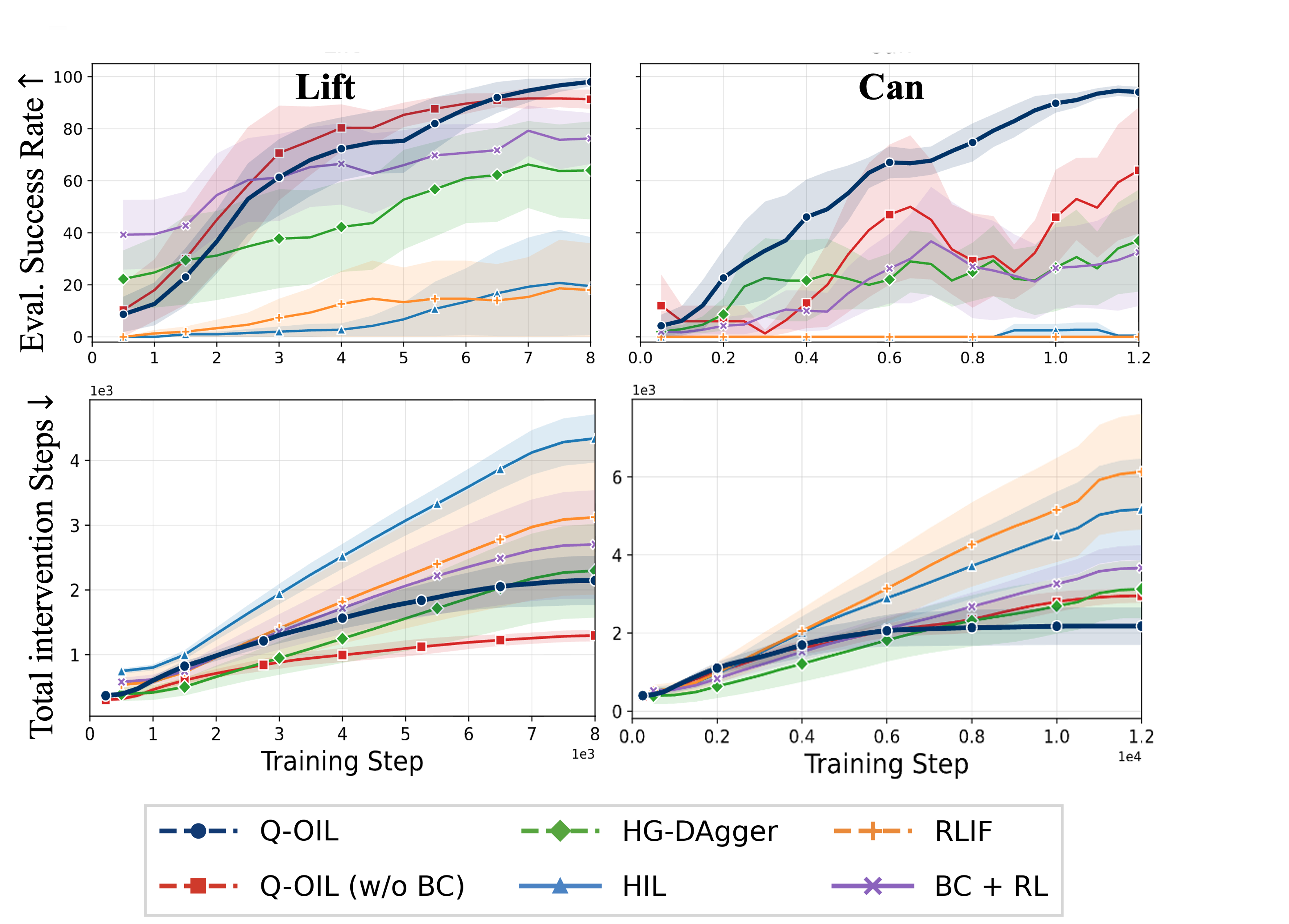}
    \caption{\footnotesize \Acronym{} outperforms baselines when human users interact with simulated robots across two tasks.}
    \label{fig:sim_real_humans}
\end{wrapfigure}
Figure~\ref{fig:sim_real_humans} (top), we observe that \Acronym{} outperforms all baselines with real humans, consistent with our simulated intervention results. HG-DAgger~\citep{michael2019hg}  has non-trivial performance but over-relies on imitating the interventions (which are suboptimal), and therefore underperforms. HIL/RLIF~\citep{luo2024hilserl, luo2024rlifinteractiveimitationlearning} fail to learn within the interaction 
budget. In contrast, \Acronym{} uses optimism to explore the human-guided regions and rapidly learns a policy with a high success rate. We also find that BC regularization enables faster and lower-variance learning in the longer-horizon ($\sim400$ steps) task. We observe similar trends between real-human interventions and our synthetic intervention model, supporting the fidelity of our data-driven evaluation. Figure~\ref{fig:sim_real_humans} (bottom) shows that \Acronym{} requires fewer interventions than baselines, while HIL~\citep{luo2024hilserl} requires substantially more effort.

\textbf{Does \Acronym{} improve efficiency and performance on real robot tasks?}

\begin{wrapfigure}{r}{0.5\textwidth}
    \vspace{-0.5cm}
    \centering
        \includegraphics[width=\linewidth]{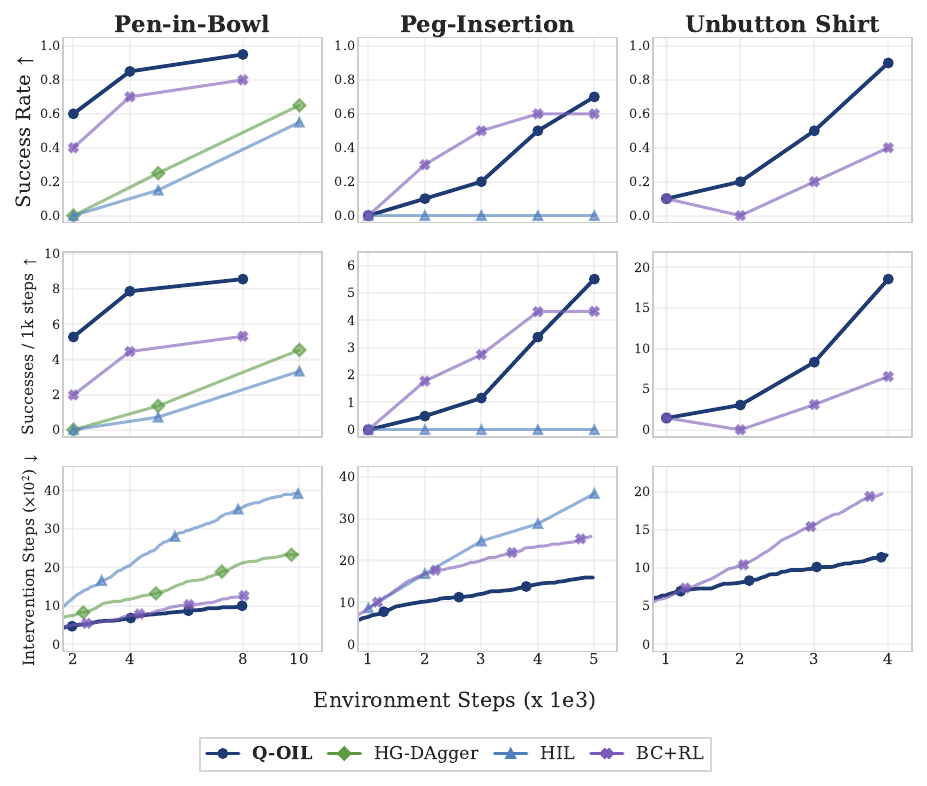}
        \caption{\footnotesize Across three tasks, \Acronym{} achieves higher success and throughput given a fixed interaction budget, with lower human effort.}
        \label{fig:real_success}
        \label{fig:real_effort}
        \vspace{-0.3cm}
\end{wrapfigure}

We validate \Acronym{} on real robot hardware using a Franka Panda (Appendix~\ref{app:robot_setup}), across three different tasks and RL baselines. Our analysis in Section~\ref{sec:human_analysis} shows that human interventions can be noisy and divergent, and \Acronym{} robustly utilizes interventions to learn a policy with $10$--$50\%$ higher success rates and almost $30\%$ lower effort (Figure~\ref{fig:real_success}). By biasing exploration toward human-corrected regions, \Acronym{} collects more task-relevant experience within the same interaction budget and learns policies with an average $90\%$ higher execution efficiency, as reflected in higher throughput, measured as successes per 1k steps. We include complete videos of our training runs, and some qualitative discussion on our website~\footnote{\url{https://weirdlabuw.github.io/qoil}}. Finally, in Figure~\ref{fig:real_success}, we observe that \Acronym{} is the most efficient in terms of interventions across tasks, i.e., it makes maximal use of each human correction to adapt its behavior. Together with our simulation results under synthetic and real corrections, these results show that human-grounded evaluation leads to better algorithmic design that transfers to real users and robots. HIL fails to learn within a smaller budget, so, for a task we run baselines for additional steps to get non-trivial task successes.

\textbf{Does \Acronym{} scale to improve pretrained VLAs on bimanual tasks?}

\begin{figure*}[h!]
    \centering
    \begin{subfigure}[b]{\linewidth}
        \centering
        \includegraphics[width=\linewidth]{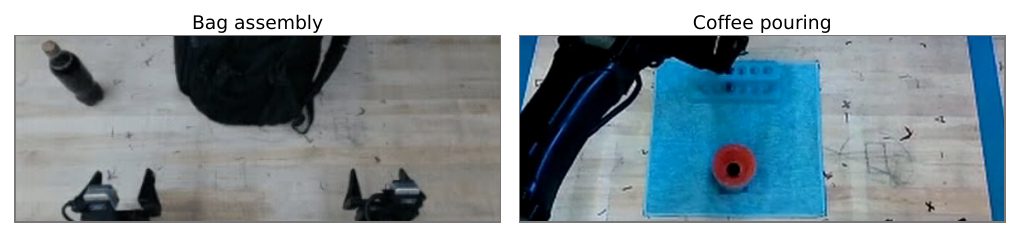}
        \vspace{-0.5cm}
        \caption{}
    \end{subfigure}
    \begin{subfigure}[b]{0.66\linewidth}
        \centering
        \includegraphics[width=\linewidth]{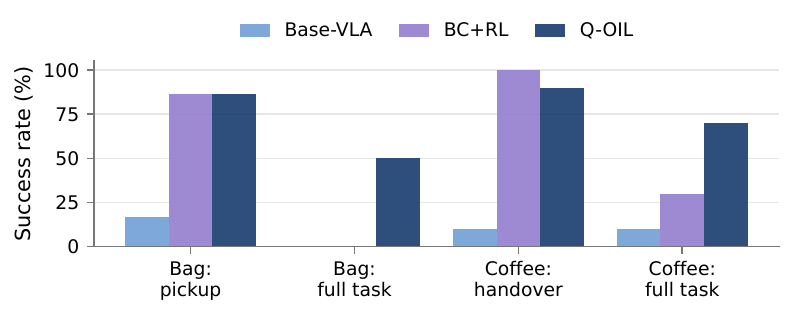}
        \caption{}
    \end{subfigure}\hfill
    \begin{subfigure}[b]{0.30\linewidth}
        \centering
        \includegraphics[width=\linewidth]{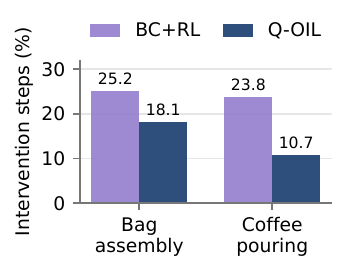}
        \caption{}
    \end{subfigure}
    \vspace{-0.2cm}
    \caption{\footnotesize
    \textbf{Bimanual VLA Fine-Tuning.}
    (a) Task setups: bag assembly and coffee pouring.
    (b) Success rates of the base VLA, BC+RL, and \Acronym{}. Bag assembly reports bottle pickup and full task completion over 30 evaluation episodes; coffee pouring reports handover and full task completion over 10 episodes.
    (c) Fraction of training steps under human intervention.
    }
    \label{fig:bimanual_vla}
    \vspace{-0.35cm}
\end{figure*}

We evaluate whether \Acronym{} can improve a pretrained VLA on two challenging bimanual tasks: bag assembly and coffee pouring (Figure~\ref{fig:bimanual_vla}). We compare residual \Acronym{} against the base VLA and the best-performing baseline BC+RL, using EXPO-FT~\citep{dong2026expo} for residual policy learning. On both tasks, \Acronym{} and BC+RL improve substantially over the base VLA on the first subtask, i.e., picking up the bottle or handing over the test tube. However, \Acronym{} completes the full task far more often ($50\%$ versus $0\%$ on bag assembly and $70\%$ versus $30\%$ on coffee pouring) while requiring fewer human interventions. We hypothesize that \Acronym{} guides local exploration to correct the most challenging subtask, i.e., stretching the side pocket to insert the bottle or pouring the beans into the beaker, leading to higher success rates with a comparable training interaction budget.

\textbf{Are the decoupled critics essential for \Acronym{} stability?}

\begin{wrapfigure}{r}{0.4\textwidth}
    \vspace{-1cm}
    \centering
    \includegraphics[width=\linewidth]{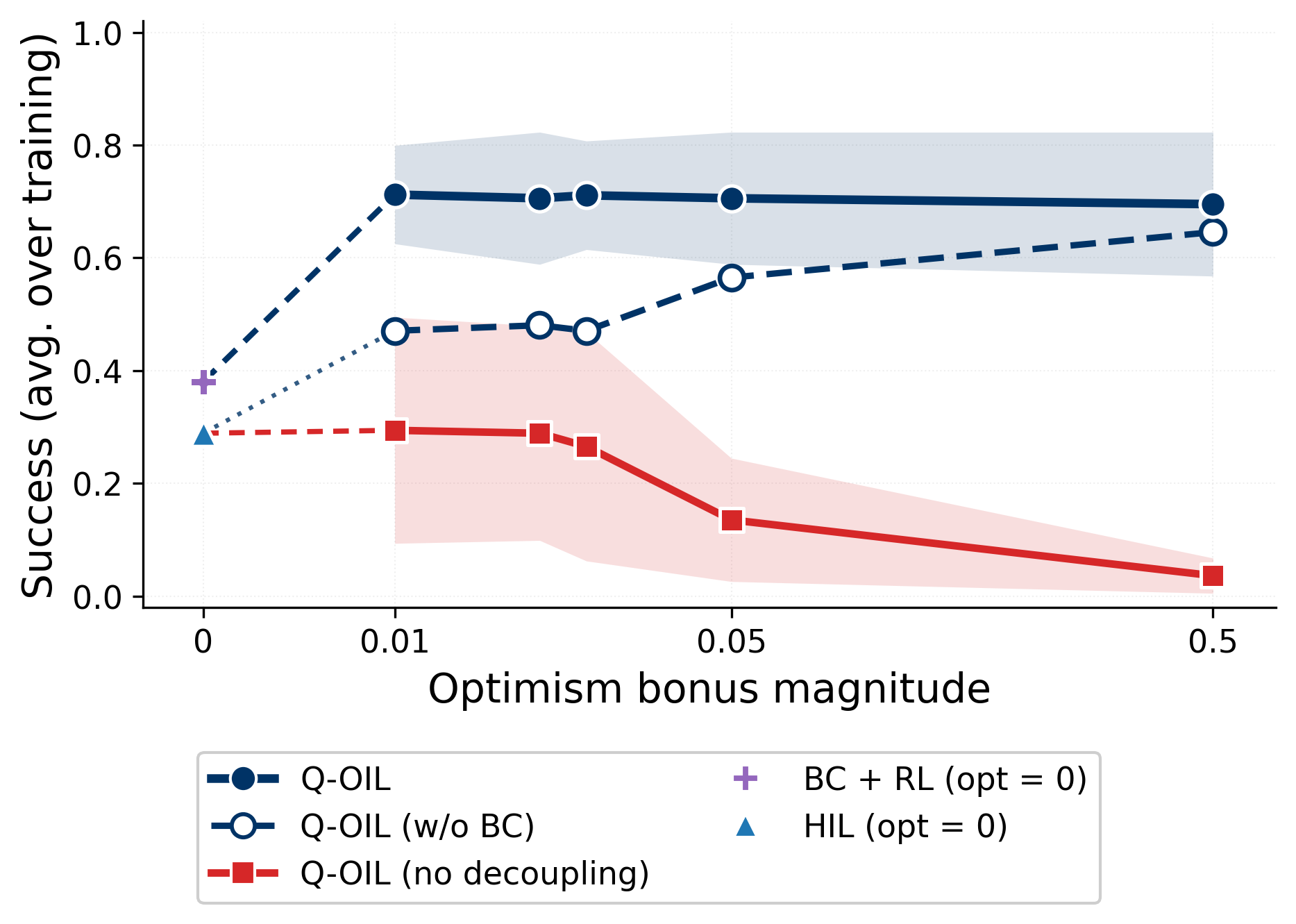}
    \caption{\footnotesize \Acronym{} w/o decoupling degrades with higher bonus due to over-optimism.}
    \label{fig:decouple_ablation}
    \vspace{-0.4cm}
\end{wrapfigure}

In Figure~\ref{fig:decouple_ablation}, we ablate a key design choice: removing the decoupling between the task reward and optimism bonus. We present the success rate averaged over Robomimic tasks with our synthetic human model. We sweep the optimism bonus $b\in[0.01,0.5]$ for \Acronym{} with/without BC and decoupling, and show that \Acronym{}'s success is stable across this $50\times$ range. At $b{=}0$, \Acronym{} reduces exactly to BC+RL. No-decoupling fails at high $b$. In Appendix~\ref{app:toy_decoupled_q}, we show a toy example in which removing decoupling leads to catastrophic propagation of positive bonuses to non-intervention states, causing over-optimism and higher values for suboptimal states. This severely harms performance, as shown in Figure~\ref{fig:decouple_ablation}, where performance decreases as we increase the bonus value for optimism.

\vspace{-0.2cm}
\section{Conclusion}

In this work, we study real-world human interventions, revealing a significant gap between existing assumptions and actual behavior. Analyzing real intervention data, we find that humans intervene based on non-Markovian progress rather than local suboptimality, and that their corrections are suboptimal and diverge from the converged policy. These findings motivate two contributions: (1) a progress-aware intervention model (PAM) that better supports algorithm development and evaluation by matching real-user outcomes, unlike prior synthetic human models, and (2) \Acronym{}, an algorithm that uses interventions to guide exploration by adding an optimism bonus at intervention transitions, while decoupling the critics for backups and policy improvement. Across simulated and real robot experiments, \Acronym{} improves task success by approximately $10$--$50\%$ with $30\%$ less human effort than baselines, providing a practical path toward scalable human-in-the-loop learning. 

\emph{Limitations:} Our human study uses 8 users in simulated tasks across 24 experimental runs; a larger and more diverse participant pool in the real world would further strengthen our findings about interventions. We evaluate \Acronym{} on tabletop manipulation tasks; future work should study robots deployed in unstructured settings, where humans intervene across diverse, long-horizon tasks with richer intervention patterns. During training, our proposed method upweights intervention transitions, which updates the policy to explore these regions. Function approximation propagates it to similar transitions, shaping exploration beyond intervention states. An interesting future direction is to leverage foundation models to improve bonus propagation beyond local exploration to a semantically richer space such as language-based interventions.

\clearpage

\bibliography{references}

\clearpage
\appendix
\startcontents[appendices]

\section*{Appendix Contents}
\printcontents[appendices]{}{1}{\setcounter{tocdepth}{2}}

\clearpage

\section{Do decoupled critics help when learning from interventions?}\label{app:toy_decoupled_q}

\begin{wrapfigure}{r}{0.5\textwidth}
    \centering
    \includegraphics[width=\linewidth]{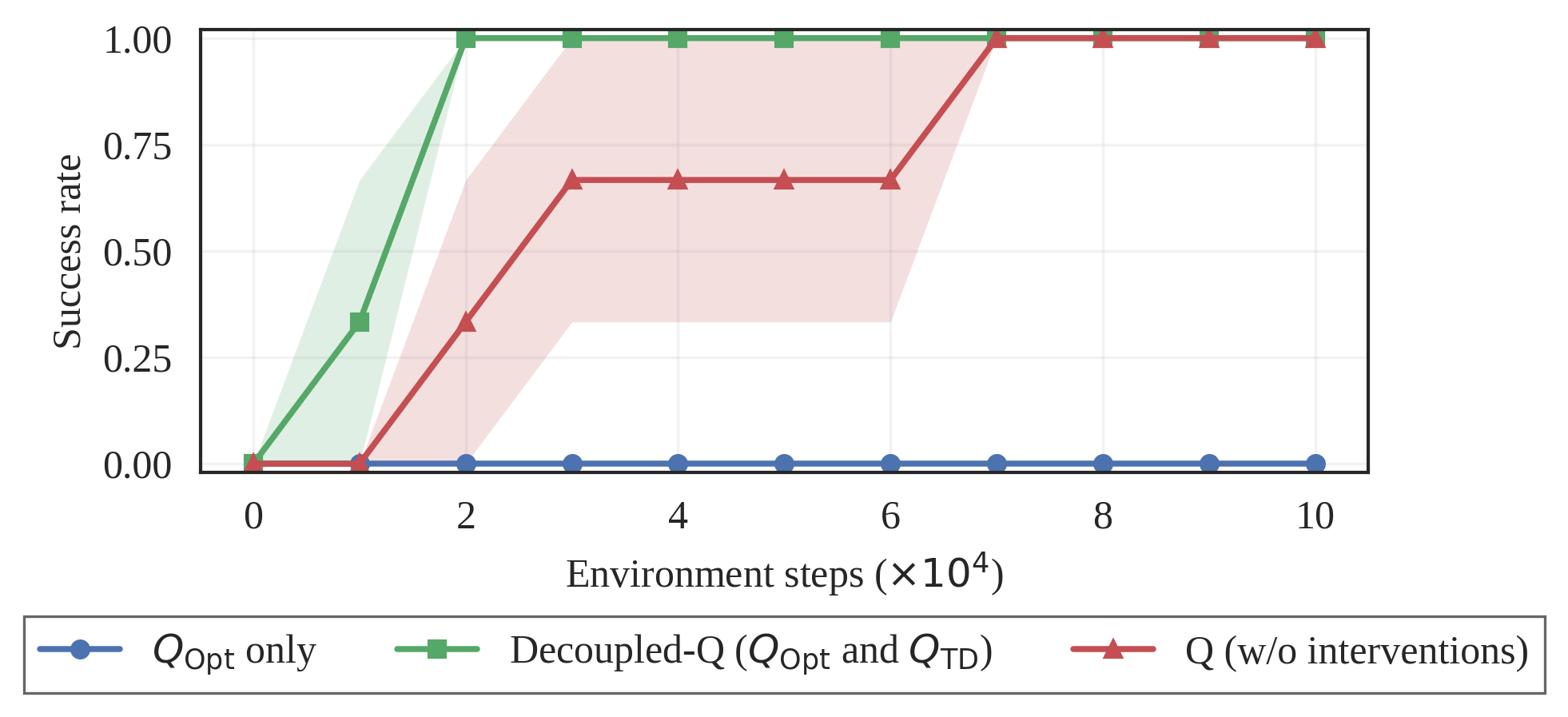}
    \caption{\Acronym{} helps learning in the discrete 2D navigation setting, showing that decoupling is essential for learning from interventions.\label{fig:maze_result}}
\end{wrapfigure}
In Section~\ref{sec:rl_method}, we introduce a key design choice in \Acronym{}: decoupling the critics into two components. The first is a task critic, which uses standard Bellman regression with only the sparse task reward. The second is an optimistic critic, which combines the task reward, the optimistic bonus, and backups from the task critic to construct Bellman targets. This design prevents the bonus from leaking to pre-intervention states. In contrast, a single critic that incorporates optimism and bootstraps from itself would propagate the bonus to states before the interventions. These are precisely the transitions where the robot was failing or acting sub-optimally, causing humans to intervene. By decoupling the critics, \Acronym{} confines the bonus to the intervention states and avoids value overestimation, which otherwise leads to catastrophic performance degradation, as shown in Figure~\ref{fig:decouple_ablation}.

\begin{figure}[h!]
    \centering
    \begin{subfigure}[b]{0.49\linewidth}
        \centering
        \includegraphics[width=\linewidth]{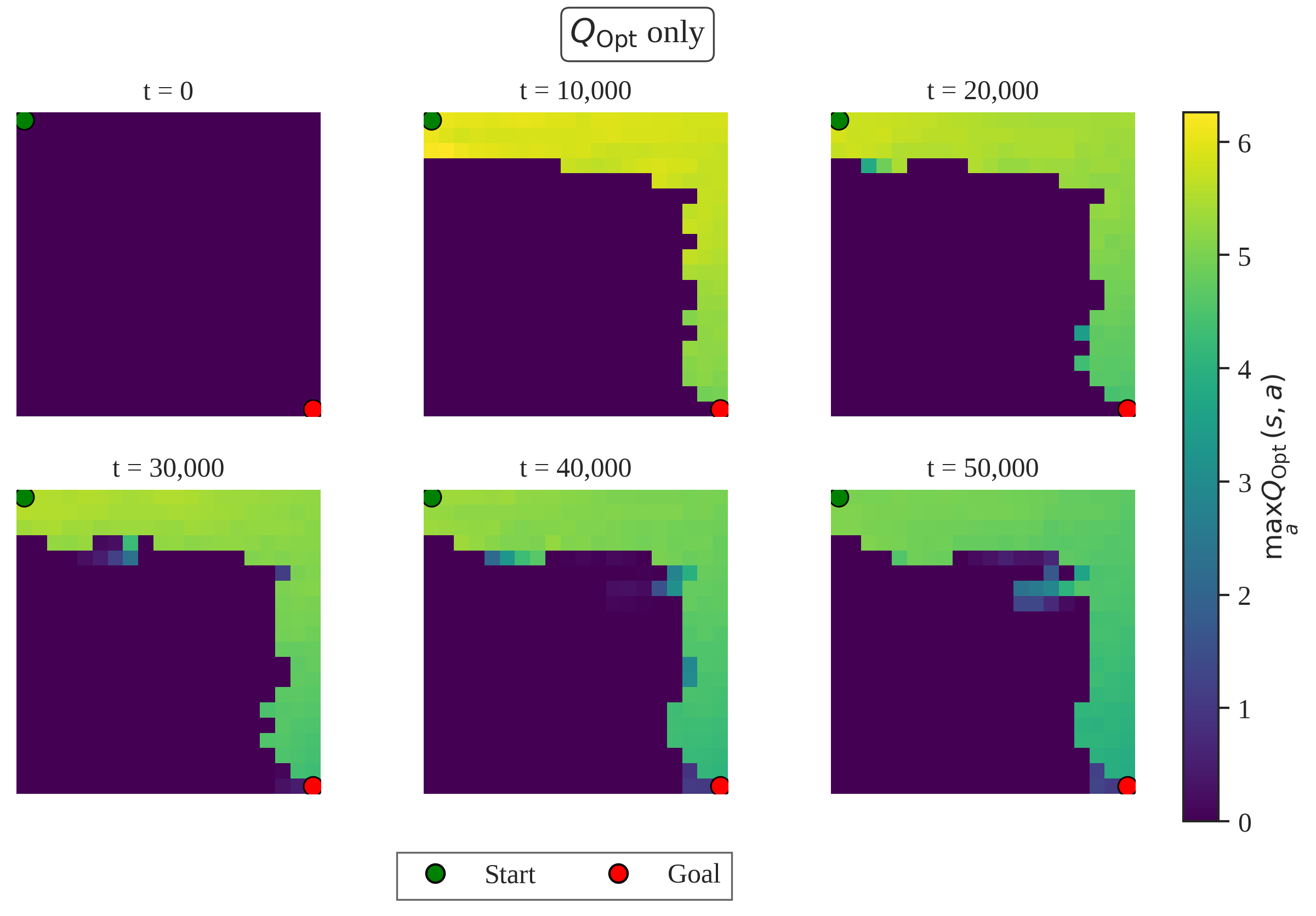}
        \caption{No Decoupling\label{fig:qopt_grid_1}}
    \end{subfigure}\hfill
    \begin{subfigure}[b]{0.49\linewidth}
        \centering
        \includegraphics[width=\linewidth]{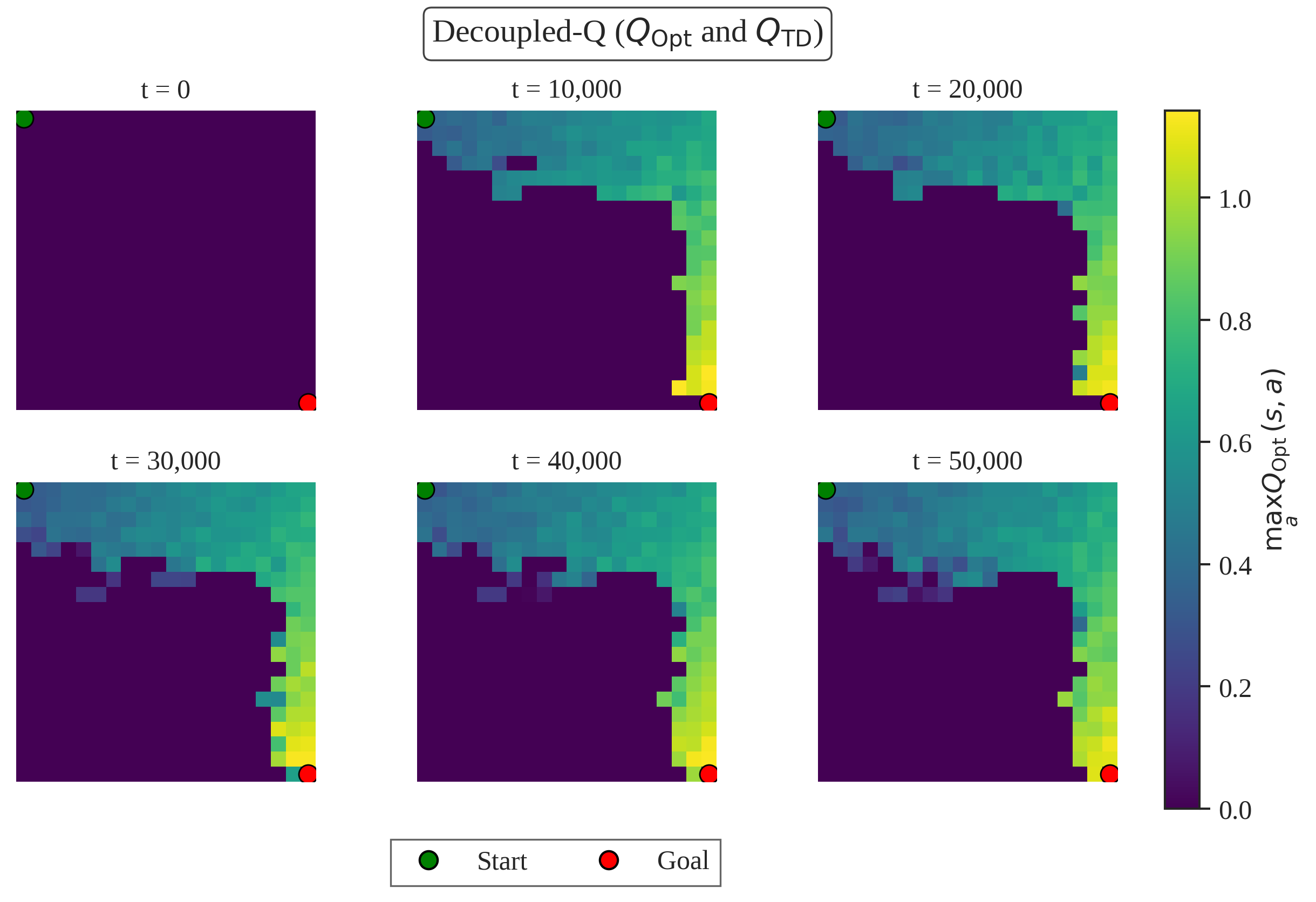}
        \caption{Decoupled Critics\label{fig:qopt_grid_2}}
    \end{subfigure}\hfill
    \caption{Visualizing the Q-value heatmaps during training across the decoupled and non-decoupled versions of \Acronym{}. Using a single critic overestimates the value because of the bonus leakage, while decoupling stabilizes Q-learning, enabling efficient learning from interventions.}
    \label{fig:qopt_grid}
\end{figure}

In Figure~\ref{fig:qopt_grid}, we present a toy example to visualize this behavior. The task is to navigate from a fixed start point to a fixed goal in a discrete $20\times20$ grid. We use a simple Q-learning algorithm to learn a policy, with a noisy expert in the loop providing interventions. As training progresses and human interventions occur in Figure~\ref{fig:qopt_grid_1}, the bonus propagates to all visited states in the grid, leading to value overestimation. The policy extracted from this overestimated value function fails to make progress toward the goal and gets stuck, as shown in Figure~\ref{fig:maze_result}. With decoupling, the Q-value heatmaps are more stable and better shaped, exhibiting a clear gradient toward the goal. This enables the extraction of an efficient policy that completes the task.

\section{Human interventions suboptimally improve robot trajectories}
\label{app:progress_metric}

To quantify action quality, we compare the value of a chosen action to the value of the expert action at the same state. For sparse goal-reaching tasks, the optimal value can be interpreted as a discounted measure of time-to-goal: if taking action $a$ reaches the goal in $n(s,a)$ steps, then $Q^*(s,a) \propto \gamma^{n(s,a)}$. Thus, the log-ratio of action values converts value differences into differences in expected steps to the goal. For an action $a$, we define
\[
\text{Subopt}(s,a)
\triangleq
1 + \frac{\log_{\gamma}\big(Q^*(s,a)/Q^*(s,a^*)\big)}{\log_{\gamma} Q^*(s,a^*)}
= 1 + \frac{n(s,a) - n(s,a^*)}{n(s,a^*)}
= \frac{n(s,a)}{n(s,a^*)}.
\]
Here, $n(s,a)$ denotes the number of steps to the goal after taking action $a$ at state $s$. Therefore, $\text{Subopt}(s,a)$ measures the relative time-to-goal induced by action $a$ compared to the optimal action $a^*$: a value of $1$ indicates an optimal action, while larger values indicate actions that delay task completion. Under this metric, intervention actions have mean suboptimality $3.16$, compared to $1.91$ for human demonstrations and $1.00$ for $\pi^*$. This suggests that intervention actions are more suboptimal than offline demonstrations under $Q^*$, while the learned expert policy is optimal by construction.
\begin{wrapfigure}{r}{0.38\textwidth}
      \vspace{-8pt}
      \centering
      \includegraphics[width=0.36\textwidth]{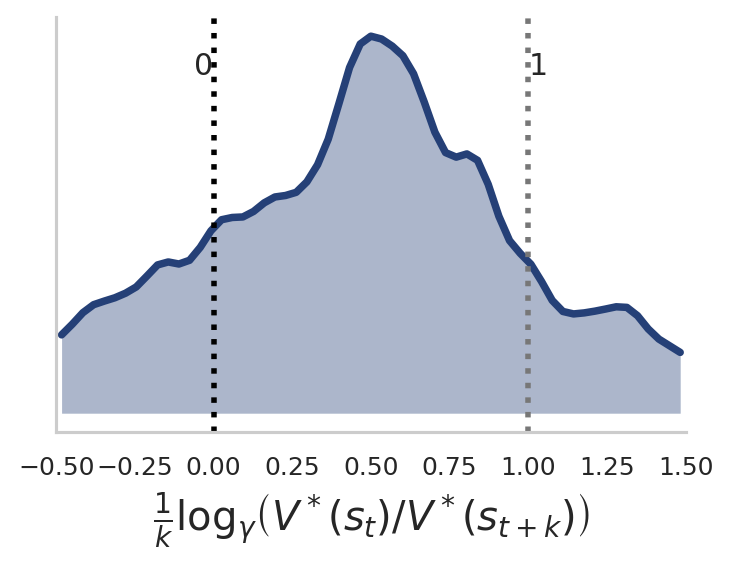}
      \caption{\footnotesize Human intervention segments often make limited value-based progress.}
      \label{fig:intervention_progress_hist}
      \vspace{-7pt}
\end{wrapfigure}
We present some additional analysis to understand how interventions affect progress toward the goal. To measure whether a takeover moves the robot closer to success, we score each intervention segment from $s_t$ to $s_{t+k}$ using,
\[
\mathrm{Progress}(s_t \!\to\! s_{t+k})
=
\frac{1}{k}\log_\gamma\!\left(
\frac{V^*(s_t)}{V^*(s_{t+k})}
\right),
\]
where $k$ is the segment length. Under sparse terminal rewards, $V^*(s)\approx \gamma^{d^*(s)}$, so this score approximates the reduction in optimal steps-to-go per environment step. A score of $1$ indicates near-expert progress, values between $0$ and $1$ indicate slower, but positive, progress, and negative values indicate negative progress. We found that $83.0\%$ of intervention segments fall below $1$, and $30.1\%$ are negative, showing that human takeovers frequently fail to make expert-like progress.

\section{Simulated Robot Interface}\label{app:sim_robot}
\begin{wrapfigure}{r}{0.5\textwidth}
    \centering
    \includegraphics[width=\linewidth]{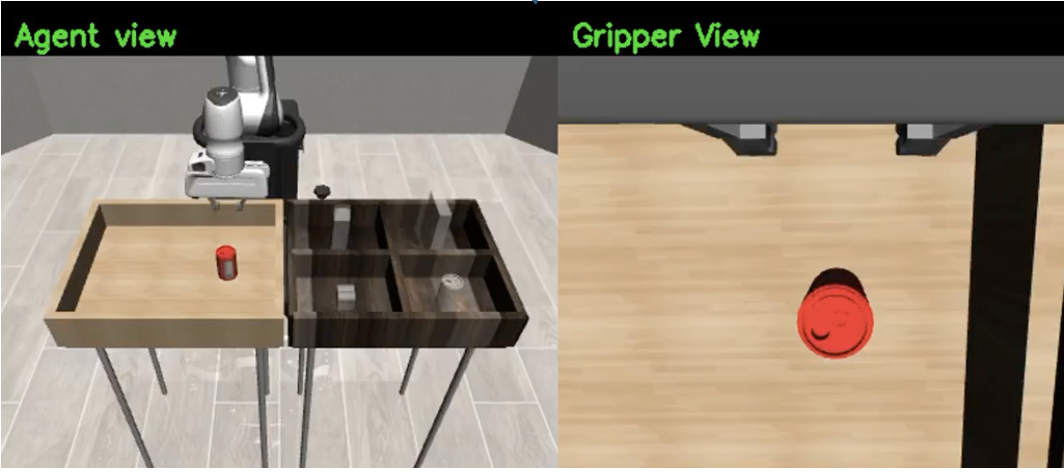}
    \caption{Rendered observation shown to the humans during simulated robot experiments.}
    \label{fig:simintvframe}
\end{wrapfigure}
We conduct these experiments to analyze human interventions with simulated robots because it enables faster and more scalable data collection and, more importantly, we can train reference policies/critics for controlled quantitative analysis. Participants teleoperated the simulated robot end-effector using a 3Dconnexion SpaceMouse. The interface displayed two synchronized rendered camera views, shown in Figure~\ref{fig:simintvframe}: an external agent view showing the robot, workspace, objects, and target receptacle, and a close-up gripper view centered on the end-effector and manipulated object. During each episode, participants monitored these views and could intervene by taking control with the SpaceMouse. SpaceMouse inputs were mapped to Cartesian end-effector position deltas, while a binary command controlled gripper opening and closing. We restricted the teleoperation interface to these 4-DoF commands, rather than full pose control, to reduce the difficulty of providing timely and consistent interventions.

\section{Human Experiment Details}\label{app:human_exps}

\subsection{Real Intervention Analysis}\label{app:human_exps_analysis}

To study how humans intervene during robot learning, we recruited eight participants familiar with robot teleoperation. Our institutional review board approved the study. We first onboarded participants to the intervention protocol and the simulated robot interface described in Appendix~\ref{app:sim_robot}. We conducted the study in simulation to enable faster and more scalable data collection, and to obtain converged reference behavior for controlled quantitative analysis.

Each participant supervised the robot on three robomimic tasks while it was trained with HIL~\citep{luo2024hilserl}. During training, participants observed the robot and intervened when they judged its behavior was unsuccessful or making insufficient progress. To limit variation and speed up convergence, we kept the starting end-effector pose and the goal position constant for each task. Across all participants and tasks, we logged the full robot trajectories, including autonomous actions, human intervention actions, and the final trained policies. Using these logs together with the task-specific reference functions described in Section~\ref{sec:human_analysis}, we analyze real human interventions. Our results show that humans tend to intervene when robot progress stalls, and that their corrective actions are often suboptimal and diverge from the converged policy.

\begin{figure}[h!]
  \centering
  \includegraphics[width=\linewidth]{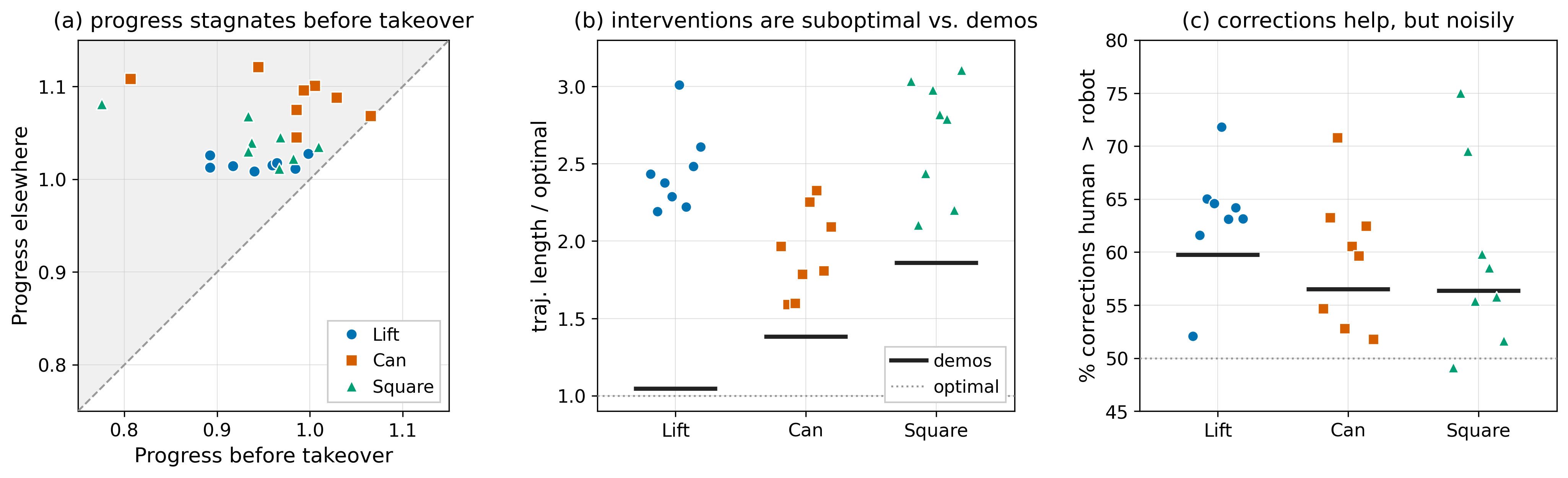}
    \vspace{-0.5cm}
    \caption{\footnotesize Per-participant results. The study has 8 participants and $\sim$2k trajectories ($\sim$200k transitions) across 3 tasks; the findings hold for every participant and task. (Finding~1) Progress before takeovers is lower than elsewhere, i.e., humans intervene as the robot stalls. (Finding~2) Relative to optimal trajectory length, interventions are costlier than demonstrations for every participant. (Corrections) The human action has higher $Q^*$ than the robot action in 51.8--69.9\% of corrections, i.e., corrections help but are noisy, motivating their use as soft guidance. \label{tab:human_study}}
    \vspace{-0.3cm}
\end{figure}

\subsection{Comparing algorithm performance across synthetic and real interventions}\label{app:human_exps_method}

In Section~\ref{sec:human_analysis}, we observe that human supervisors intervene when they perceive the robot's task progress to be insufficient and that the corrections they provide are suboptimal. We next assess whether our progress-aware synthetic intervention model can act as a reliable proxy for real human supervision when benchmarking robot learning algorithms. To obtain a real-human reference, four participants supervised online training of six methods, \Acronym{}, \Acronym{} w/o BC, BC+RL, HG-DAgger, HIL~\citep{luo2024hilserl} and RLIF, on the Lift and Can tasks with randomized robot initial positions. We trained each method for 8,000 environment steps on Lift and 12,000 environment steps on Can. We report the resulting success rates across training checkpoints in Figure~\ref{fig:sim_real_humans} and rank the methods by their final success rates in Figure~\ref{fig:ranks}. We replicate the evaluation with three synthetic intervention models: our progress-based gating model, an intervention model based on value/action suboptimality, and a random intervention model tuned to match the observed human intervention frequency. For each synthetic model, we run every learning algorithm with multiple random seeds and present the resulting success-rate curves in Figure~\ref{fig:intv_sim_results}. For every intervention source, we order the learning algorithms by their final success rate and then evaluate how well the synthetic orderings correlate with the real-human ordering using Spearman rank correlation. As shown in Figure~\ref{fig:correlation}, our progress-based model attains perfect rank correlation with the real-human results. In other words, it reproduces the same algorithm ranking as real users, suggesting it more faithfully maintains the algorithm-level takeaways one would obtain from actual human-in-the-loop studies.

\section{Real-World Setup}\label{app:robot_setup}

Our real-world setup follows the HIL-SERL system of \citet{luo2024hilserl}. We use a robot manipulator equipped with a parallel-jaw gripper and train vision-based policies directly on the physical robot. RGB cameras placed around the workspace provide visual observations. For each task, the robot policy uses a wrist camera for observations. For tasks that require broader scene context, we also use an external side-view camera. The observations also include proprioceptive information, including the current end-effector state and gripper state.

\begin{wrapfigure}{r}{0.4\textwidth}
    \vspace{-0.5cm}
    \centering
    \includegraphics[width=\linewidth]{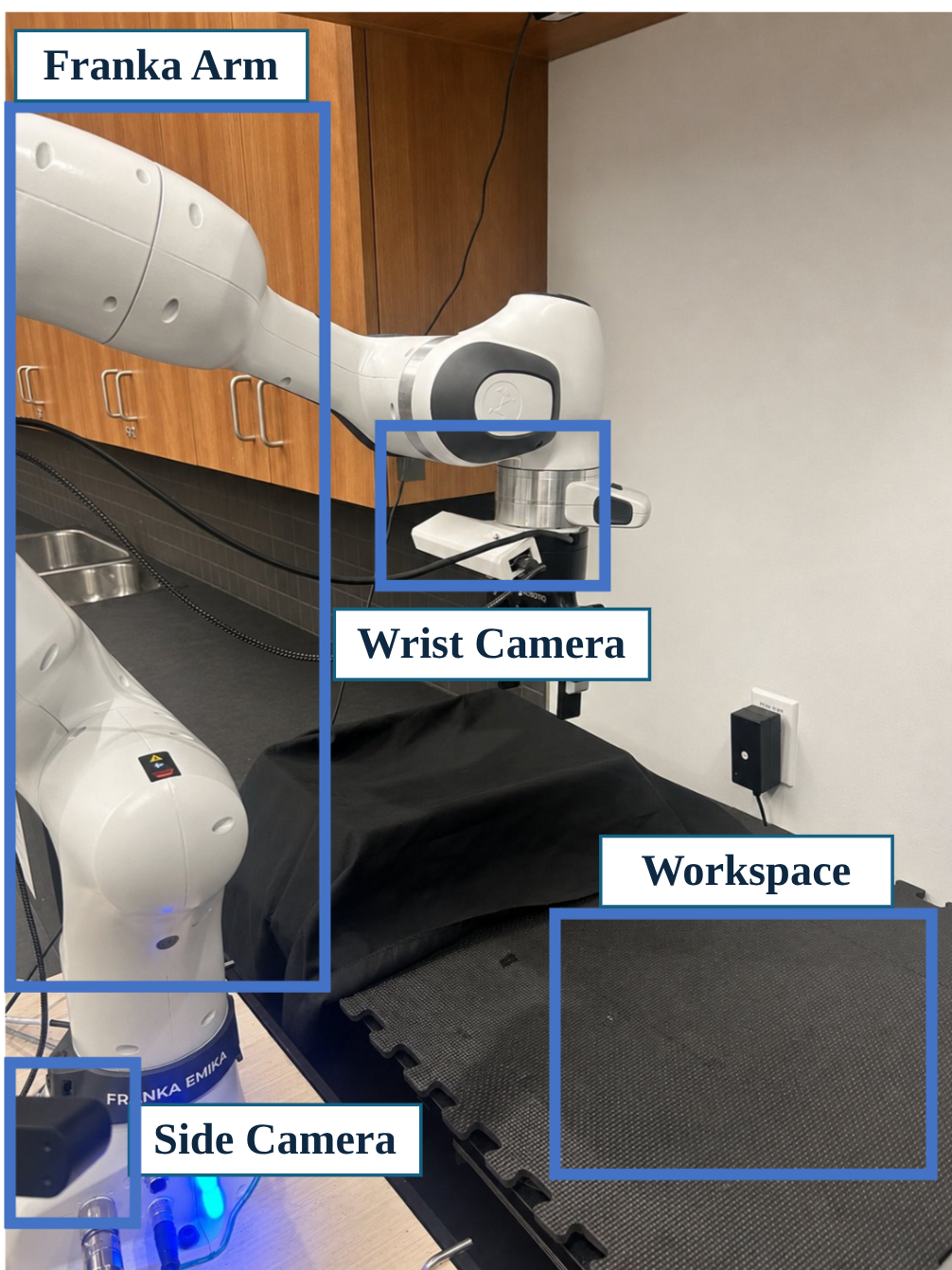}
    \caption{\footnotesize Real-world robot setup. The policy receives RGB observations from a wrist-mounted camera and, optionally, a side-view camera. The human supervisor monitors the robot during online training and provides interventions using a SpaceMouse.}
    \label{fig:real_robot_setup}
\end{wrapfigure}

Following \citet{luo2024hilserl}, we implement a hybrid action space: a Gaussian policy outputs 3-DoF Cartesian end-effector position deltas $(x,y,z)$, while a separate discrete critic selects the binary gripper command. We remove end-effector rotations to simplify teleoperation, and clip all commanded motions to a safety bounding box. We also synchronize the learner and actor loops to achieve a UTD ratio of $10$ across all methods.

For each task, we first collect a small set of SpaceMouse demonstrations to initialize the demonstration replay buffer. During online training, the learned policy controls the robot by default. At the same time, the supervisor can intervene with the SpaceMouse when the robot is likely to fail, becomes stuck, or stops making progress. Intervention actions replace the policy actions on the robot and are logged for policy updates. The supervisor also manually resets each episode by randomizing the relevant objects and robot configuration within valid task ranges. We train each real-world task for up to 10k environment steps and evaluate checkpoints on a fixed set of reset conditions. The tasks are described in Appendix~\ref{app:tasks}. \textbf{We include all videos of training and evaluation on our project website: \url{https://weirdlabuw.github.io/qoil}.}

\subsection{Extending \Acronym{} to Pretrained VLAs}
\label{app:vla_setup}

\begin{algorithm}
    \caption{\footnotesize Residual \Acronym{} for Pretrained VLAs}
    \label{alg:vla_qoil}
    \begin{algorithmic}[1]
    \setlength{\itemsep}{0pt}
    \setlength{\parskip}{0pt}
    \REQUIRE Frozen VLA $\pi_{\mathrm{base}}$, residual policy $\pi_\theta$, human $\pi^{\mathrm{int}}$, buffers $D_\pi,D_{\mathrm{int}}$, bonus $b$, chunk sizes $H_b,H_r$

    \FOR{trial $i=1$ to $N$}
        \STATE \texttt{\textcolor{NavyBlue}{// Online rollout}}
        \FOR{timestep $t=1$ to $T$ in steps of $H_r$}
            \IF{$t$ starts a new base chunk}
                \STATE $A^{\mathrm{base}} \leftarrow \pi_{\mathrm{base}}(o_t)$
            \ENDIF
            \STATE $\Delta A_t \leftarrow \pi_\theta(o_t,A_t^{\mathrm{base}})$
            \STATE $\hat A_t \leftarrow A_t^{\mathrm{base}}+\eta\Delta A_t$
            \STATE Execute $\hat A_t$ for $H_r$ steps, allowing human takeover
            \FOR{transition $(o_s,a_s,r_s,o_{s+1})$ in chunk}
                \IF{$\pi^{\mathrm{int}}$ intervened at $s$}
                    \STATE $r_s^b \leftarrow b$
                    \STATE Add $(o_s,a_s^{\mathrm{int}})$ to $D_{\mathrm{int}}$
                \ELSE
                    \STATE $r_s^b \leftarrow 0$
                \ENDIF
                \STATE Add $(o_s,a_s,r_s,r_s^b,o_{s+1})$ to $D_\pi$
            \ENDFOR
        \ENDFOR
        \STATE \texttt{\textcolor{NavyBlue}{// Decoupled critic updates}}
        \STATE Train $Q_{\mathrm{TD}}$ on $D_\pi$ via $\mathcal{L}_{\mathrm{TD}}$
        \STATE Train $Q_{\mathrm{Opt}}$ on $D_\pi$ via $\mathcal{L}_{\mathrm{Opt}}$
        \STATE \texttt{\textcolor{NavyBlue}{// Residual policy improvement}}
        \STATE Update $\pi_\theta$ using $Q_{\mathrm{Opt}}$ and BC on $D_{\mathrm{int}}$
    \ENDFOR
    \end{algorithmic}
\end{algorithm}

\textbf{Task and Robot Setup.}
We evaluate our method on two bimanual tasks. In bag assembly, the robot must use two arms to pick up a bottle, stretch the side pocket of a backpack, and insert the bottle accurately. In coffee pouring, which mimics a wet-lab task, the robot must pick a test tube filled with coffee beans from a rack and pour the beans into a beaker. We use the I2RT YAM bimanual robot with a leader-follower setup, where the leader arms mirror the follower arms during autonomous learning, allowing the human supervisor to seamlessly take over the leaders and intervene when needed. We use joint-space control and RGB observations from an external scene camera. We first collect a small set of human demonstrations to fine-tune the pretrained VLA, then freeze it during online learning.

\textbf{Online Training.}
During online training, the robot executes actions from the pretrained VLA with learned residual corrections by default. The human supervisor can take over the leader arms when the robot gets stuck or fails to make progress, and these corrective actions update the residual policy online. We manually reset the bottle and backpack between episodes.

\textbf{Residual Policy Learning.}
For each task, we first fine-tune a pretrained VLA on 25 human demonstrations and use the resulting policy as a frozen base policy. Following EXPO~\citep{dong2026expo}, we learn a residual policy that corrects the VLA's predicted actions. The base VLA predicts action chunks of length $H_b=32$, and the residual policy predicts corrections over shorter chunks of length $H_r=8$. For each base action, the residual policy is queried multiple times using updated RGB observations, allowing the policy to make closed-loop corrections at a higher frequency without repeatedly querying the base VLA. We compose the executed residual action chunk as
\begin{equation}
    A_t = A_t^{\mathrm{base}} + \eta \pi_\theta^{\mathrm{res}} (o_t,A_t^{\mathrm{base}}),
\end{equation}
where $A_t^{\mathrm{base}}$ is the corresponding $H_r$ actions from the base VLA and $\eta$ controls the magnitude of the residual actions. We apply per-dimension scaling to account for the different limits of joint and gripper actions. This enables the residual policy to locally improve the pretrained VLA's behavior without updating its parameters.

\textbf{Learning from Interventions.}
We use residual SAC with BC regularization on intervention actions as our baseline. To extend \Acronym{}, we maintain two critics over the executed actions: a task critic $Q_{\mathrm{TD}}$ trained on the environment reward and an optimistic critic $Q_{\mathrm{Opt}}$ trained with an additional positive reward on intervention transitions. Importantly, the optimistic critic bootstraps from the task critic rather than itself, preventing the intervention bonus from propagating to preceding states. We update the residual policy to maximize $Q_{\mathrm{Opt}}$, with an auxiliary BC loss on human corrections. The base VLA remains frozen throughout online training. We use a UTD ratio of $2$ for both \Acronym{} and BC+RL. Algorithm~\ref{alg:vla_qoil} summarizes this procedure.

\textbf{Evaluation.}
For bag assembly, we evaluate the learned policies on three sets of 10 episodes each and report both bottle pickup and full task completion rates. For coffee pouring, we evaluate on 10 episodes and report handover success and full task success, where at least half of the coffee beans are poured into the beaker. We additionally measure human effort during online adaptation as the fraction of control steps under intervention. For bag assembly, the BC+RL and \Acronym{} training runs include 36 and 17 takeovers, corresponding to intervention fractions of $25.2\%$ and $18.1\%$, respectively; for coffee pouring, the intervention fractions are $23.8\%$ and $10.7\%$.

\section{Synthetic Intervention Models and Rank-Correlation Analysis}
\label{app:intervention_models}
\label{app:correlation}

\subsection{Parameter Selection for the Progress-Aware Model}

Our synthetic model predicts whether an intervention should occur based on recent task progress. Its two main parameters are the window length $k$, which sets the time horizon over which we assess progress, and the threshold $\delta$, which specifies when we consider progress stalled; we set the intervention probabilities $\alpha,\beta$ based on empirical intervention rates observed in the human data. We tune these parameters using held-out human intervention data. Concretely, we run a grid search over $k \in \{1, 2, 3, \dots, 32\}$ and
$\delta \in [-0.5, 0.5]$, and choose the setting that maximizes F1 score for binary intervention prediction. We optimize F1 over accuracy because interventions are sparse: a model can obtain high accuracy by predicting not to intervene always, but such a model is not useful for simulating human supervision. The F1 objective balances precision and recall, encouraging the model to identify true intervention states without triggering excessively. We fit a single set of parameters across all tasks and participants, rather than tuning separate parameters per task, so that the resulting model captures general task-agnostic intervention behavior and can transfer to new tasks. We then fix the selected parameters for all simulated intervention experiments.

\subsection{Held-Out Intervention Prediction}

Table~\ref{tab:prediction} reports intervention prediction performance on held-out human data. The random model is calibrated to the empirical human intervention rate. The action-suboptimality model triggers interventions based on whether the robot action is suboptimal under a learned value estimate. The progress-aware model achieves the best precision and recall, indicating that recent lack of task progress is a better predictor of when humans intervene than either random triggering or instantaneous action suboptimality.

\begin{table}[!ht]
    \centering
    \caption{Intervention prediction on held-out human data.}
    \label{tab:prediction}
    \begin{tabular}{lcc}
        \toprule
        \textbf{Method} & \textbf{Precision} & \textbf{Recall} \\
        \midrule
        Action Suboptimality & 0.1136 & 0.4489 \\
        Random & 0.0837 & 0.0837 \\
        \textbf{PAM (Ours)} & \textbf{0.1750} & \textbf{0.6059} \\
        \bottomrule
    \end{tabular}
\end{table}

\subsection{Rank Correlation with Real-Human Evaluation}

\begin{wrapfigure}{r}{0.5\textwidth}
    \vspace{-0.5cm}
    \centering
    \includegraphics[width=\linewidth]{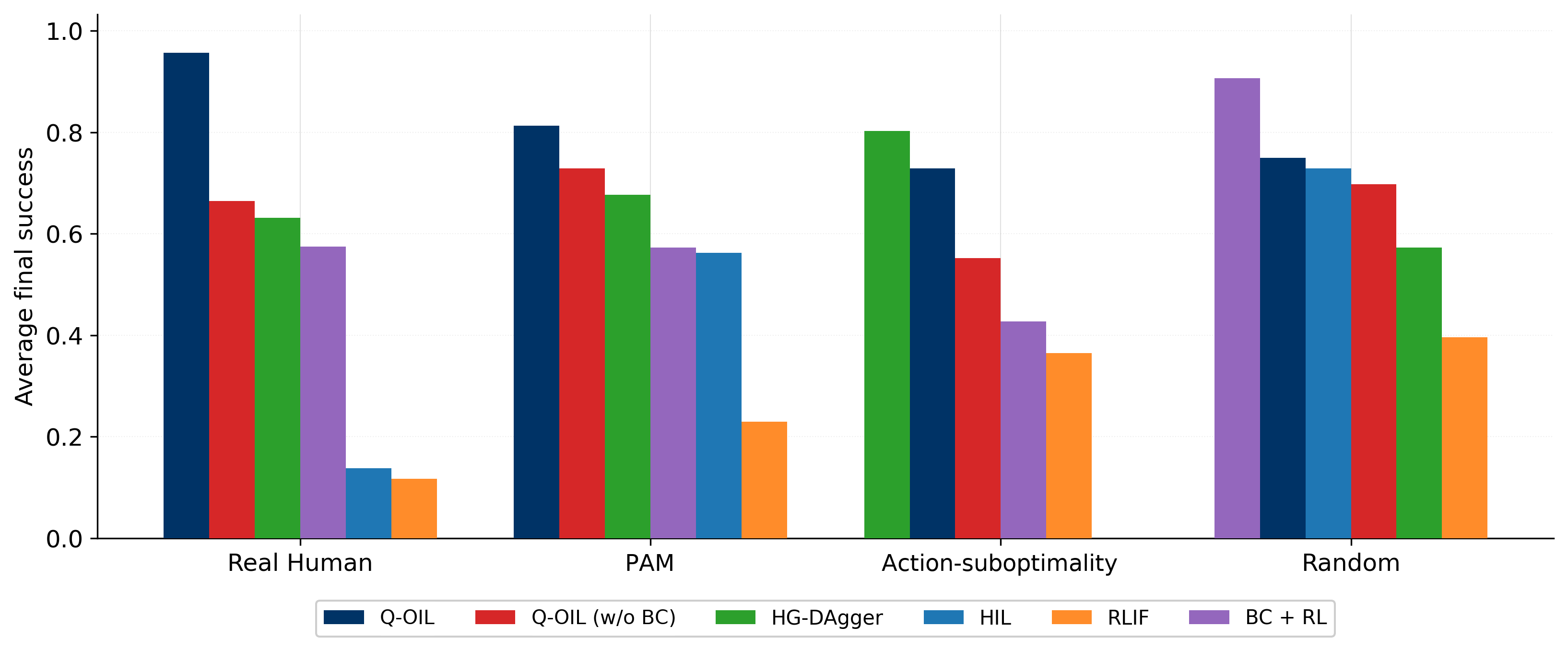}
    \caption{\small Rankings of algorithms under real human evaluation and three simulated intervention models, based on final success rate.}
    \label{fig:ranks}
\end{wrapfigure}

We next evaluate whether synthetic intervention models preserve the same algorithm-level conclusions as real human supervisors. To obtain the real-human reference ranking, participants supervised online training of the evaluated algorithms as discussed in Appendix~\ref{app:human_exps_method}. We ranked the algorithms by final success rate. Figure~\ref{fig:ranks} shows the rankings used for the Spearman rank-correlation analysis in Figure~\ref{fig:correlation}. A Spearman correlation of $\rho=1.0$ means that the synthetic intervention model produces the same ordering of algorithms as real-human supervision, even if the absolute success rates differ.

\subsection{Synthetic Intervention Rollouts}

In each synthetic evaluation, the intervention model determines when the simulated supervisor intervenes. Once an intervention is triggered, the simulated supervisor provides corrective control for a sampled intervention duration. We sample these durations from the lower 75\% of the empirical human intervention-length distribution, and actions are generated from a suboptimal RL checkpoint with $\sim$50--70\% success rate matching empirical success rates from real human intervention experiments. This design choice ensures simulated interventions are suboptimal (matching our analysis in Section~\ref{sec:human_analysis}) while still providing useful directional guidance.

\subsection{Algorithm Performance under Different Synthetic Models of Human Intervention}

\begin{wrapfigure}{r}{0.4\textwidth}
    \centering
    \includegraphics[width=\linewidth]{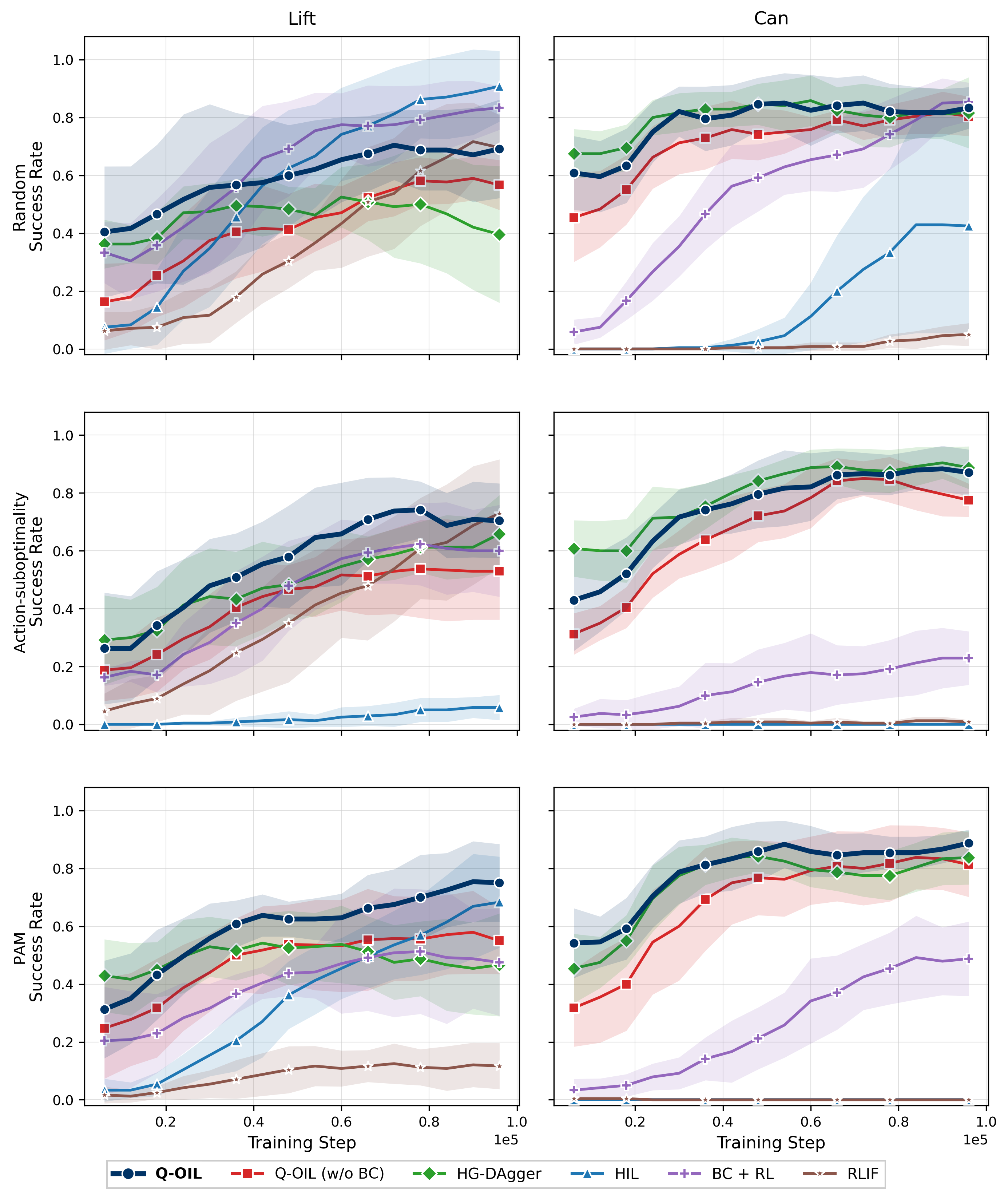}

    \caption{\footnotesize Success rate vs.\ env steps under different intervention models}
    \label{fig:intv_sim_results}
    \vspace{-1.5cm}
\end{wrapfigure}

We evaluate all learning algorithms under the synthetic intervention models described in Section~\ref{sec:rethinking}. Figure~\ref{fig:intv_sim_results} shows success rates for the PAM, Random, and Action-suboptimality models. For each intervention model, we run each learning algorithm across multiple random seeds and report success rate as a function of environment steps. The results show that the choice of intervention model can substantially affect conclusions about baseline performance. In particular, the relative ordering of the
baselines changes across intervention models: HIL improves under Random interventions but performs worse under Action-suboptimality, while HG-DAgger is favored by the Action-suboptimality model. In contrast, PAM matches the real-human ranking (from both simulated and real robot tasks), suggesting that modeling when humans intervene based on stalled progress better preserves the conclusions obtained from human-in-the-loop experiments.

\section{Task Details}\label{app:tasks}

\begin{figure}[h!]
    \centering
    \includegraphics[width=0.6\linewidth]{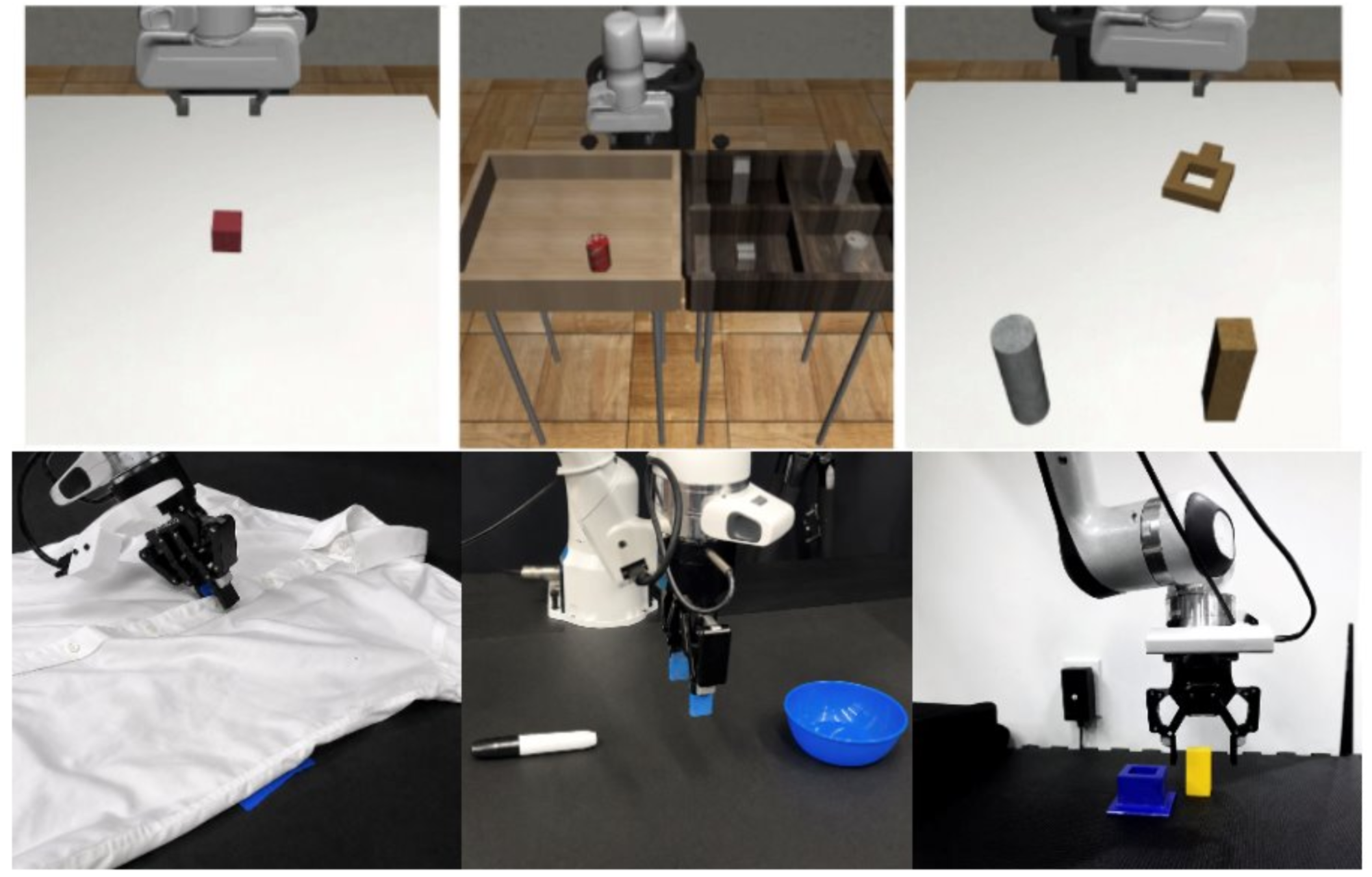}
    \caption{Simulated Robomimic tasks (top) and Franka tasks (bottom). The bimanual YAM tasks are shown in Figure~\ref{fig:bimanual_vla}.}
    \label{fig:tasks}
\end{figure}

We evaluate \Acronym{} on a set of real-world and simulated manipulation tasks that require learning from sparse rewards with human interventions (see Figure~\ref{fig:tasks}).

\paragraph{Pen-in-Bowl.}
A Franka arm must pick up a pen and drop it into a fixed bowl using side- and wrist-camera observations. At the start of each episode, the pen pose is randomized within a bounded region. The task uses a sparse reward, $r=1$, when the pen is detected inside the bowl, with the reward learned using a classifier as in~\citet{luo2024hilserl}. Episodes have a horizon of 200 environment steps. We evaluate on 20 fixed pen positions. This task benefits from higher interaction throughput: when the robot initially fails to place the pen in the bowl, it can often recover by regrasping the pen and completing the task within the same episode.

\paragraph{Unbutton Shirt.}
A robot must manipulate a deformable shirt placed on a table and open a button using wrist-camera observations. To simplify grasping, we fix the roll, pitch, and yaw of the end effector. The task uses a sparse success reward, $r=1$, upon successfully unbuttoning the shirt. We evaluate over 10 episodes.

\paragraph{Peg Insertion.}
A robot must insert a peg into its corresponding hole in a 3-DoF setup using wrist-camera observations. The initial state is randomized at reset, and the task uses a sparse success reward. We evaluate on 10 fixed test positions.

\paragraph{Bimanual YAM Tasks.} The bag assembly and coffee pouring tasks are described in Appendix~\ref{app:vla_setup}.

\paragraph{Robomimic Simulation.}
We also evaluate on the Can, Square, and Lift tasks from Robomimic. Simulation experiments use a 7-DoF robot, while human-intervention experiments use a 4-DoF control interface. Initial states are randomized over the default environment ranges. Policies are evaluated over 100 episodes, and simulation results with synthetic intervention models are reported over three random seeds.

\section{How does quality of expert actions affect learning?}
\label{app:perfect_experts}
\begin{figure}[h!]
    \centering
    \includegraphics[width=\linewidth]{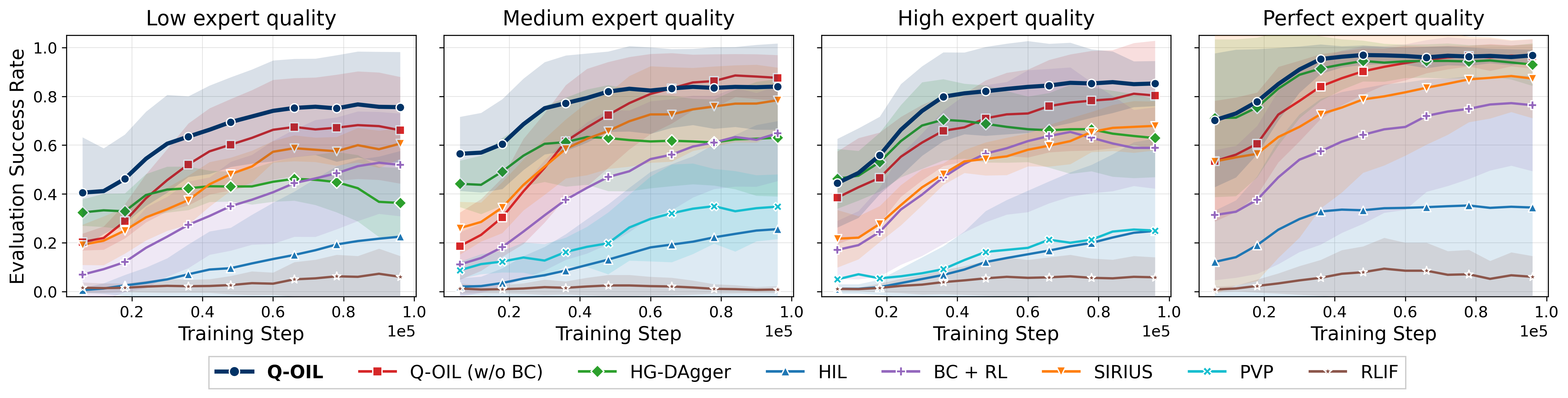}
    \caption{\small Success across takeover-quality tiers, averaged over Lift, Can, and Square (3 seeds). Tiers group intervention checkpoints by per-task success rate (Low: 42/44/58\%, Medium: 60/80/72\%, High: 68/94/82\%, Perfect: 100\%).}
    \label{fig:expert_ablation}
\end{figure}

In Figure~\ref{fig:expert_ablation}, we ablate the quality of the corrective actions by grouping intervention policies into four tiers by success rate. With perfect experts (100\% success rate), most methods converge to a high success rate efficiently, and imitation-based baselines (e.g., HG-DAgger) are competitive as they receive near-optimal corrections. As expert quality decreases, \Acronym{} (with or without BC) remains the best-performing method, with the largest margins under the weakest experts, while other baselines degrade as they inefficiently use noisy and suboptimal corrections. This aligns with our observations across experiments with real humans and real robots. Overall, \Acronym{} depends on intervention quality, but less than the baselines, making it practical for real-world deployment where intervention quality varies.

\section{Implementation Details}
\label{app:implementation}

\subsection{Hyperparameters}
All methods use the same demonstrations, reward definitions, actor–critic architecture, replay buffer size, evaluation episodes, and intervention interface unless otherwise stated.

\begin{table}[!ht]
    \centering
    \caption{Hyperparameters for \Acronym{} and baselines.}
    \label{tab:hyperparams}
    \begin{tabular}{ll}
        \hline
        \textbf{Parameter} & \textbf{Value} \\
        \hline
        \multicolumn{2}{l}{\textit{Core RL Parameters}} \\
        Architecture & MLP with Gaussian Head \\
        Hidden layers & 3 layers of width 256 \\
        Optimizer & Adam \\
        Learning rate & 3e-4 \\
        Discount $\gamma$ & 0.99 \\
        Soft update $\tau$ & 0.005 \\
        UTD Ratio & 10 \\
        Batch size & 256 \\
        Replay buffer size & 1e6 \\
        Learnable temperature & False \\
        \hline
        \multicolumn{2}{l}{\textit{\Acronym-Specific}} \\
        Optimism bonus $b$ & 0.025 \\
        Critic ensemble size (each of TD / Opt) & 10 \\
        BC regularization weight & 0.1 \\
        \hline
        \multicolumn{2}{l}{\textit{Intervention Model}} \\
        Stagnation window $k$ & 9 \\
        Stagnation threshold $\delta$ & -0.158 \\
        Intervention probability $\alpha$ & 0.599 \\
        Intervention probability (no stagnation) $\beta$ & 0 \\
        \hline
    \end{tabular}
\end{table}

\begin{table}[!ht]
    \centering
    \caption{Hyperparameters for residual VLA learning with \Acronym{} and BC+RL.}
    \label{tab:residual_hyperparams}
    \begin{tabular}{ll}
        \hline
        \textbf{Parameter} & \textbf{Value} \\
        \hline
        \multicolumn{2}{l}{\textit{Base Policy and Residual Architecture}} \\
        Base policy & Pretrained VLA (frozen) \\
        Base policy demonstrations & 25 \\
        Base policy fine-tuning steps & 20k \\
        Residual policy & Gaussian MLP + ResNet-10 \\
        Actor / critic hidden layers & 2 layers of width 256 \\
        Action space & Bimanual joint space (14-DoF) \\
        Base action chunk size $H_b$ & 32 \\
        Residual action chunk size $H_r$ & 8 \\
        \hline
        \multicolumn{2}{l}{\textit{Residual Policy Parameters}} \\
        Residual joint scale & 0.02 rad \\
        Residual gripper scale & 0.10 \\
        Initial edit scale & 0.25 \\
        Final edit scale & 1.0 \\
        Edit scale step size & 0.05 \\
        \hline
        \multicolumn{2}{l}{\textit{Core RL Parameters}} \\
        Algorithm & Residual SAC~\cite{dong2026expo} \\
        Optimizer & Adam \\
        Learning rate & 3e-4 \\
        Discount per control step & 0.97 \\
        Soft update $\tau$ & 0.005 \\
        UTD Ratio & 2 \\
        Batch size & 256 \\
        Critic ensemble size & 2 \\
        BC regularization weight & 0.1 \\
        Initial temperature & 0.01 \\
        Training starts (transitions) & 200 \\
        \hline
        \multicolumn{2}{l}{\textit{\Acronym-Specific}} \\
        Optimism bonus $b$ & 0.1 \\
        Additional optimistic critic & Yes \\
        Optimistic critic bootstrap & Task critic \\
        \hline
    \end{tabular}
\end{table}

\end{document}